\documentclass[journal]{IEEEtran}
\usepackage{graphicx}
\usepackage[cmex10]{amsmath}
\usepackage{array}
\usepackage[hyphens]{url}

\ifCLASSINFOpdf
\else
\fi
\usepackage{multirow}
\usepackage{array}
\usepackage{amsmath,amsfonts,amssymb}
\usepackage{mathtools}
\usepackage{xcolor}
\DeclarePairedDelimiter{\ceil}{\lceil}{\rceil}
\newcolumntype{P}[1]{>{\centering\arraybackslash}p{#1}}
\newcolumntype{M}[1]{>{\centering\arraybackslash}m{#1}}
\begin{document}
\title{ADEPOS: A Novel Approximate Computing Framework for Anomaly Detection Systems and its Implementation in 65nm CMOS}
%
%
%

\author{Sumon Kumar Bose,~\IEEEmembership{Student~Member,~IEEE,}
       Bapi Kar,~\IEEEmembership{Member,~IEEE,} Mohendra Roy,~\IEEEmembership{Member,~IEEE,} Pradeep Kumar Gopalakrishnan,~\IEEEmembership{Senior~Member,~IEEE}, Zhang Lei,    Aakash Patil,~\IEEEmembership{Member,~IEEE,} and Arindam Basu, ~\IEEEmembership{Senior~Member,~IEEE}

\thanks{This work was supported by Delta Electronics Inc. and the National Research Foundation Singapore under the Corp Lab @University scheme.}
\thanks{S. K. Bose, B. Kar, P. K. Gopalakrishnan and A. Basu are with the Delta-NTU Corporate Laboratory for Cyber-Physical Systems, School of EEE, NTU, Singapore 639798.  (e-mail: bose0003@e.ntu.edu.sg; arindam.basu@ntu.edu.sg).}
\thanks{M. Roy is with PDPU, Gandhinagar, India 382007 (e-mail: mohendra.roy@sot.pdpu.ac.in).}
\thanks{Z. Lei is with the Centre of Excellence in IC
Design (VIRTUS), School of EEE, NTU, Singapore 639798 (e-mail: CharlesZhang@ntu.edu.sg).}

\thanks{A. Patil is associated with Tork Motors, Pune, India 411026.}
}
\maketitle
\begin{abstract}
To overcome the energy and bandwidth limitations of traditional IoT systems, ``edge computing" or information extraction at the sensor node has become popular. However, now it is important to create very low energy information extraction or pattern recognition systems. In this paper, we present an approximate computing method to reduce the computation energy of a specific type of IoT system used for anomaly detection (e.g. in predictive maintenance, epileptic seizure detection, etc). Termed as Anomaly Detection Based Power Savings (ADEPOS), our proposed method uses low precision computing and low complexity neural networks at the beginning when it is easy to distinguish healthy data. However, on the detection of anomalies, the complexity of the network and computing precision are adaptively increased for accurate predictions. We show that ensemble approaches are well suited for adaptively changing network size. To validate our proposed scheme, a chip has been fabricated in UMC $65$nm process that includes an MSP430 microprocessor along with an on-chip switching mode DC-DC converter for dynamic voltage and frequency scaling. Using NASA bearing dataset for machine health monitoring, we show that using ADEPOS we can achieve $8.95$X saving of energy along the lifetime without losing any detection accuracy. The energy savings are obtained by reducing the execution time of the neural network on the microprocessor.
\end{abstract}


%
\IEEEpeerreviewmaketitle

\section{Introduction}

%
%
%
%
\IEEEPARstart{A}{nomaly} detection can be stated as the identification or separation of unusual and abnormal data distribution from the normal one. Early detection of anomaly alerts us about the unusual neuronal activities in patients with an epileptic seizure, the preliminary stage of machine breakdown~\cite{7727444} or the credit card fraud and network intrusion. In this paper, we will focus on anomaly detection in the context of predictive maintenance of the machine and epileptic seizure.

Machine breakdown laggards the throughput and productivity of industries eventually leading to losses in revenue. Periodic maintenance of machines reduces the downtime due to unexpected machine failure but causes over maintenance. On the contrary, predictive maintenance (PdM) keeps a balance between unnecessary maintenance and run to failure of machines. Typically vibrations~\cite{Qiu2006}, \cite{Girdhar2004}, temperature~\cite{Porotsky2012}, humidity, etc. signals are sensed to monitor the machine health condition in the PdM approach. Due to the slow dynamics of temperature or humidity signal, it can be sampled once every second and the amount of generated data is quite low. However, the sampling rate is much higher for detecting machine vibrations. Even though the vibration signal is acquired in a duty cycled approach, most of the energy and bandwidth have to be allocated to send a large chunk of vibration data from the sensing node to the cloud. This conventional way of data processing poses two issues in battery-powered sensor nodes: 1) excessive drainage of battery energy which limits the lifetime of sensor nodes 2) latency to get the decision of processed data from cloud servers. These issues can be resolved by edge computing~\cite{7879243},~\cite{Basu2018} framework where the data processing tasks are offloaded from the cloud to IoT nodes.

In this paper, we present an edge computing framework which reduces the power consumption of the sensor node exploiting the output of the anomaly detector by dynamic voltage and network scaling (DVNS). Dynamic voltage scaling (DVS) is one of the popular methods to achieve energy optimization in electronics system by supplying power at the optimum energy point. Wide supply range on-chip switched capacitor (SC) converter~\cite{6737288} has been implemented for DVS but it suffers from low efficiency and on-chip area overhead. 
However, for lithium or button cell battery powered IoT nodes where the area is limited by the battery volume, inductor based switching converter is suitable due to its higher efficiency and wide supply range~\cite{7030240}.

The energy requirement of the system can be brought down further by approximate computing where slight errors in calculation either do not change the final outcome or can be ignored by human perception~\cite{Raha2015}. Approximations in the computation are introduced from the basic building block of a circuit such as an adder~\cite{Lu2004},~\cite{Improving} or multiplier~\cite{Kulkarni2011},~\cite{low_power} to the system level~\cite{Raha2017},~\cite{Raha2018xxx},~\cite{Optimum}. Since the neural network is good approximator, in this work, we have introduced approximation in the calculation at the system level using dynamic network expansion (DNS) where we have dynamically varied the number of neurons in the system based on anomaly detector decision.

Machine learning (ML) methods have been applied previously for machine health monitoring such as for fault detection of aero-engine control system sensor~\cite{Vishwanath}, and fault diagnostics of the rotary machine~\cite{SANZ2007981}. However, insufficiency of machine failure data is one of the main bottlenecks of machine health monitoring employing ML~\cite{khan_madden_2014} for different types of fault detection. Generation of failure signatures of the machine by intentionally putting the machine in the faulty condition is not cost effective since the data collection process will completely destroy the machine. Moreover, it is very unlikely that the failure signatures of different machines and different failure cases are similar. Therefore, learning from a generic dataset is not a feasible solution. However, an overabundance of healthy data from the machine can be utilized to train PdM sensor node as one class classifier (OCC) and any deviation from the healthy data can be treated as an anomaly~\cite{Roy2019}.

\begin{figure}[t]
\centering
\includegraphics[scale=0.3]{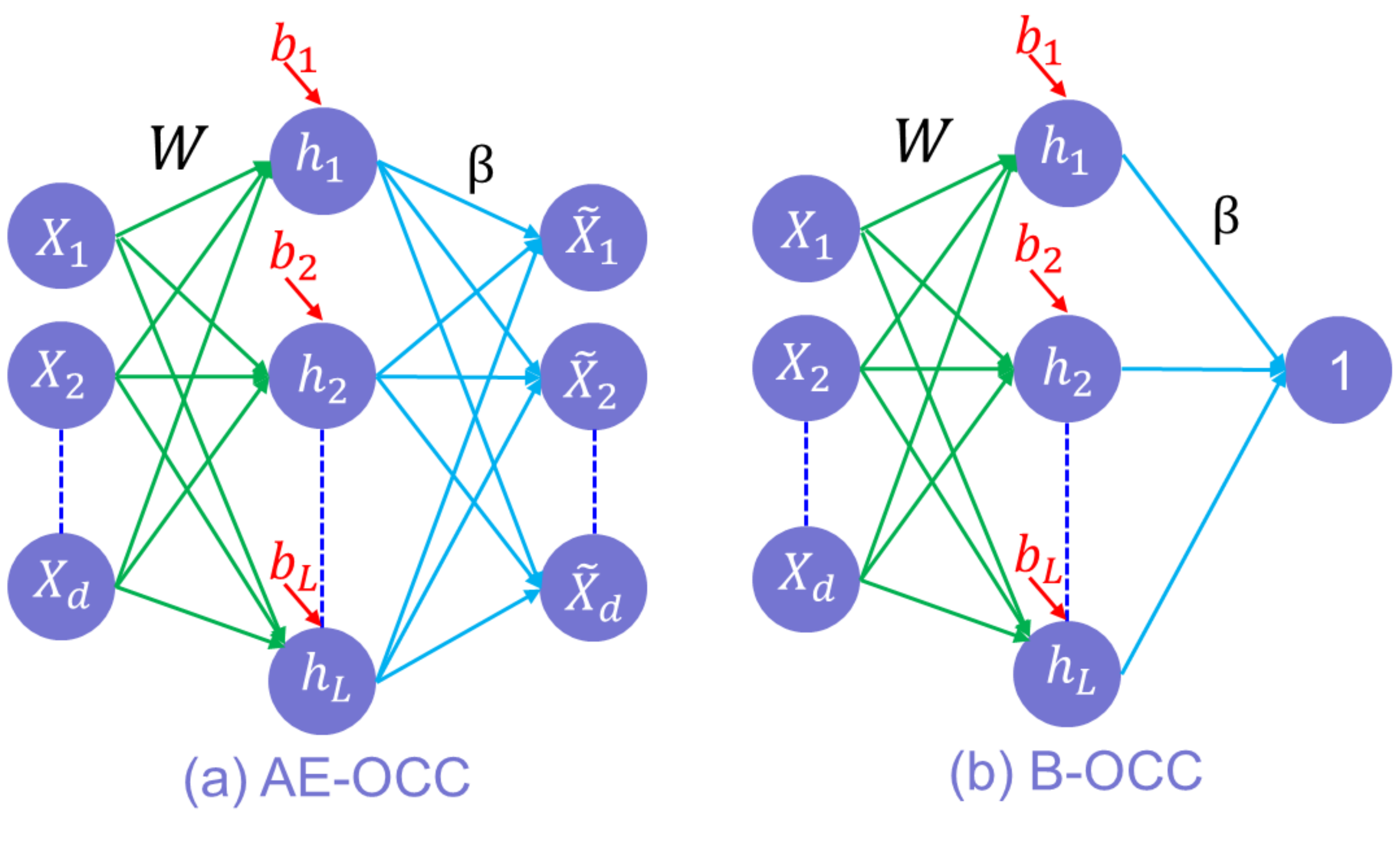}
\caption{(a) Autoencoder (AE) Architectures: In reconstruction based one class classification (OCC), AE is trained using healthy data to learn input statistics and to reconstruct the input vector at the output. (b) Boundary-based OCC is trained to produce a value of $1$ at the output for healthy data. During testing when either reconstructed output deviates from the input (reconstruction based OCC) or from $1$ (boundary based OCC) by a threshold amount, the machine can be declared as faulty.}
\label{fig1}
\vspace{-1.5em}
\end{figure}

Our major contributions in this paper are summarized as follows: 1) Application of boundary-based extreme learning machine (ELM-B) computational framework for machine health monitoring and anomaly detection 2) Adaptive usage of approximate computing (ADEPOS) along the lifetime of machine based on the anomaly detector output for low energy computation 3) Hardware demonstration of ADEPOS algorithm using dynamic voltage and network scaling (DVNS). Our proposed ADEPOS algorithm has the benefit of reducing energy in both microprocessor-based systems (this work) as well as ASIC implementations where parts of the chip can be turned OFF when unused. Compared to our earlier algorithmic work~\cite{Bose_2019}, we present hardware results from a fabricated chip for the first time that quantifies energy savings afforded by ADEPOS in actual hardware. Since most IoT systems rely on microprocessor based implementations, we demonstrate our results on such a system.

We arrange the remaining part of the paper is as follows. In section \ref{brief}, we review the basics of boundary and autoencoder based extreme learning machine (ELM-B and ELM-AE) as one class classifier (OCC) in the application of machine health monitoring whereas we present the proposed ADEPOS algorithm for energy saving in section \ref{algo}. Section \ref{hw} presents the hardware implementation for DVS and ADEPOS. In section \ref{re}, we show the results of our experiment on NASA bearing dataset. Section \ref{disc} captures the results of seizure detection using EEG dataset followed by conclusions in the last section.

\section{ELM-AE and ELM-B based OCC}
\label{brief}
\subsection{Autoencoder (AE)}
Single layer autoencoder (AE) is the simplest form of autoencoder consisting of an input layer, an output layer, and a single hidden layer. Fig. \ref{fig1}(a) captures the basic architecture of single layer autoencoder whose input weights W and biases $b=[b_1, b_2, \cdots, b_L]^T$ encode the input data distribution $X = [x_1, x_2, \cdots, x_d]^T$ into $L$ hidden neurons $h=[h_1, h_2, \cdots, h_L]^T$ whereas output weights $\beta$ decodes the latent features hidden into h to output neurons $\tilde{X} = [\tilde{x_1}, \tilde{x_2}, \cdots, \tilde{x_d}]^T$. In autoencoder, input and output dimension are same since autoencoder always attempts to reconstruct the input data at its output. Eq. (\ref{eqn1a}) and (\ref{eqn1b}) represent the basic encoding and decoding operations occurring in an autoencoder.

\begin{align}
 h_j&=g\left(\sum\limits_{i=1}^dW_{ji}x_i+b_j\right); ~~j=1,2,...,L \label{eqn1a} \\
\tilde{x_k}&=\sum\limits_{j=1}^L\beta_{jk}h_j; ~~k=1,2,...,d \label{eqn1b}\\
\tilde{R}&=\sum\limits_{j=1}^L\beta_{j}h_j; \label{eqn1c}
\end{align}

Figure \ref{fig1}(b) depicts the architecture of boundary mode single hidden layer feed-forward network (SLFN) where input features $X$ are being projected into $L$ dimensional hidden space using weights $W$ and biases $b$ and output is targeted at $1$ (or any real number $R$). Basic operations of boundary mode SLFN are highlighted in Eq. (\ref{eqn1a}) and (\ref{eqn1c}). We prefer absolute value as nonlinear activation function over sigmoid or tanh since it is easier to implement in hardware and no performance degradation was observed. 

In traditional autoencoder (TAE) or artificial neural network (ANN), optimal weights $W$, $\beta$, and biases $b$ are learned using back-propagation in an iterative way. Even though back-propagation produces state-of-the-art accuracy, it requires a large number of data samples to converge and experiences high computational overhead. On the contrary, an alternative computing paradigm based on a random high dimensional projection of input data~\cite{HUANG2006489},~\cite{Zhang2019},~\cite{Rahimi2009} provides faster convergence and needs to tune only hidden layer to output layer weights $\beta$. The other parameters, input to hidden layer weights, $W$ and biases, $b$ are taken randomly from a continuous probability distribution. In this paper, we use the term extreme learning machine (ELM) to refer to these algorithms, but any other algorithm in this class is equally applicable in our case.

In ELM framework, the optimal output layer weights $\beta^*$ in Eq. (\ref{eqn2a}) are found by minimizing error (${\epsilon}$) in the least square sense where the error (${\epsilon}$) is defined in Eq.  (\ref{eqn2anew}). $H^\dagger$ in Eq.  (\ref{eqn2a}) denotes the Moore-Penrose generalized inverse~\cite{PENROSE1954} of $H$.

\begin{align}
 \beta^*&=H^\dagger X \label{eqn2a}\\ 
 \epsilon&=
\begin{dcases}
    \Vert X-\tilde{X} \Vert ^2,& \text{if ELM-AE} \\
    \Vert R-\tilde{R} \Vert ^2,& \text{if ELM-B} \label{eqn2anew}
\end{dcases}
\end{align}
As discussed earlier, training the PdM sensors from a generic dataset and deploying on the machine is not a viable solution since the neural network model parameters are dependent on machine and sensor position and orientation. To mitigate these issues, online sequential learning of the PdM sensors deploying on the machine is necessary. ELM based online sequential learning algorithms OPIUM, OSELM have been presented in ~\cite{TAPSON201394},~\cite{4012031} respectively. However, OSELM demands matrix inverse operation which is costly in terms of memory and computes. Hence, keeping in mind the real practical application scenario of PdM sensor, we have opted for OPIUM as the learning framework for ELM based autoencoder and boundary methods since it requires less computation.

Equations for the online update of $\beta$ are shown in Eq. (\ref{eqn3a}) and (\ref{eqn3b}) using  $\theta=C\cdot I$ where $C$ is constant and $I$ is LxL identity matrix~\cite{VANSCHAIK2015233}.

\begin{align}
\eta_k&=\frac{\theta \cdot h_k}{1+h_k^T\cdot \theta \cdot h_k} \label{eqn3a} \\
\beta_k&=\beta_{k-1}+\eta_k (X_k-\beta_{k-1}h_k)\label{eqn3b}
\end{align}

\subsection{One Class classifier (OCC)}
Since the failure data from the machine is rare, we trained ELM-AE and ELM-B model as one class classifier using healthy data from the machine at the very early stage of machine life. As soon as the training of machine learning model stops, the system enters into inference mode. When the reconstructed output has error exceeding a pre-defined threshold, we can schedule the machine for maintenance.
\begin{figure}[t]
\includegraphics[scale=0.57]{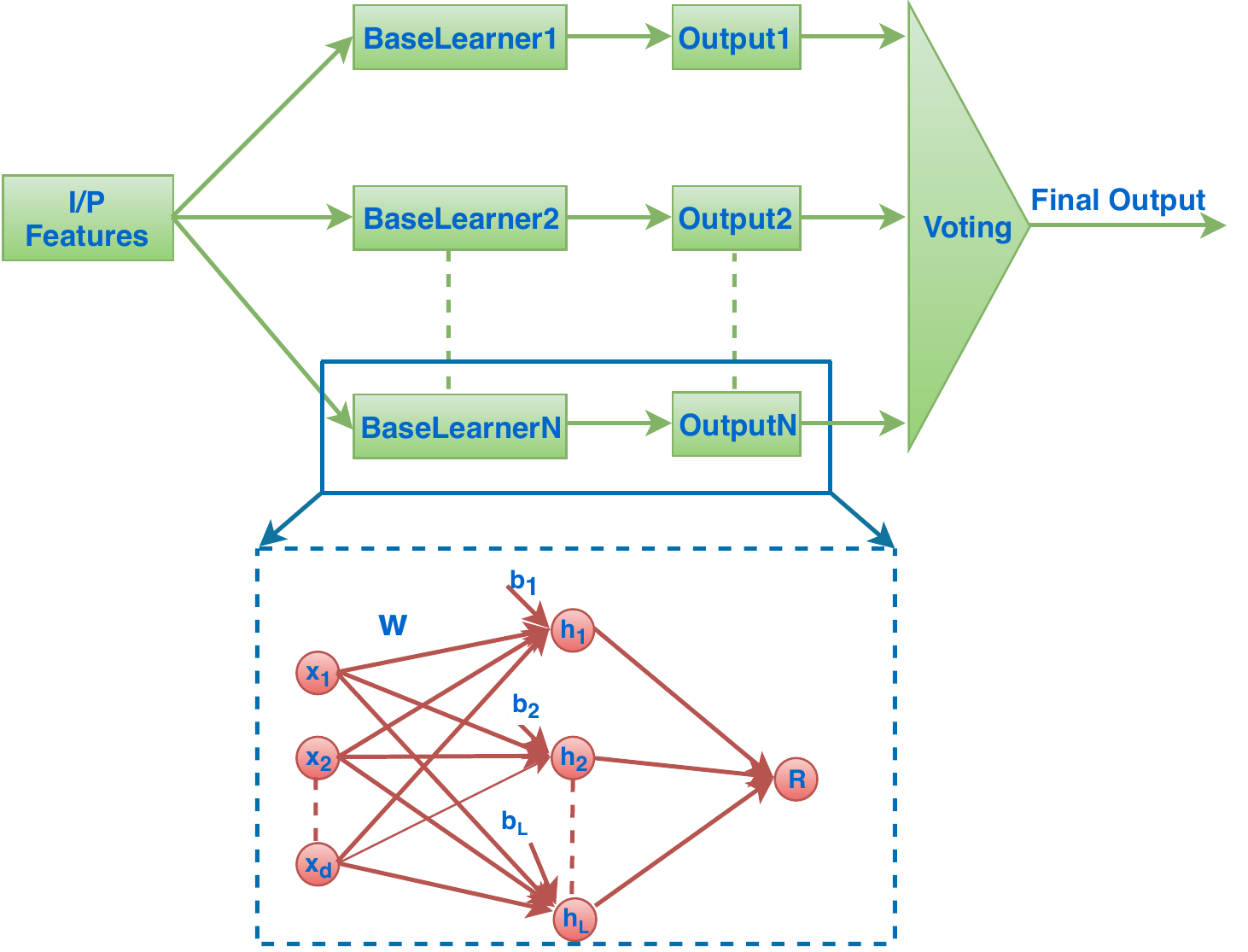}
\caption{Ensemble of $N$-base learners (BL). Each BL has $L$ hidden neurons and the final output is dependent on all intermediate BLs output on the basis of majority voting.}
\label{fig_ensm}
\end{figure}
\section{Algorithmic Improvements for Energy Saving}
\label{algo}
In this section, we will discuss two algorithmic enhancements for energy saving - a) exploiting approximate computing along the lifetime of machine based on anomaly detector output (ADEPOS) and b) generation of hidden neurons with lowest counts of multiply and accumulate (MAC) operations possible. The first approach towards energy saving is generic to any machine learning model and hardware platform whereas the second approach is only applicable to randomized first layer based computing framework.

\subsection{ADEPOS: Anomaly Detector based Power Saving}
In this paper, we proposed to use imprecise/ approximate computing \cite{6810241} along the lifetime of the machine to save energy of the anomaly detection system. At a very early stage of machine life, changes of input data distribution are minimal and anomaly detector output is far from the threshold. Hence, errors or inaccuracies introduced in the computation due to approximate computing will be lower than the margin. Once the anomaly detector raises alarm, accuracy in the computation is elevated gradually on the same input data to scrutinize whether it is truly an anomaly or a false alarm due to the approximate computing. Approximation in the calculation can be added in many ways such as by varying data path bit precision \cite{7801877} during computation, changing network size or reducing accuracy in feature extraction part, the voltage over scaling \cite{5763154} and so on. In this paper, we only discuss the dynamic changing of effective neurons in the network (DNS) and lowering supply voltage to the near-threshold region as representative parameters tuned by ADEPOS.

\begin{figure}[t]
\includegraphics[scale=0.55]{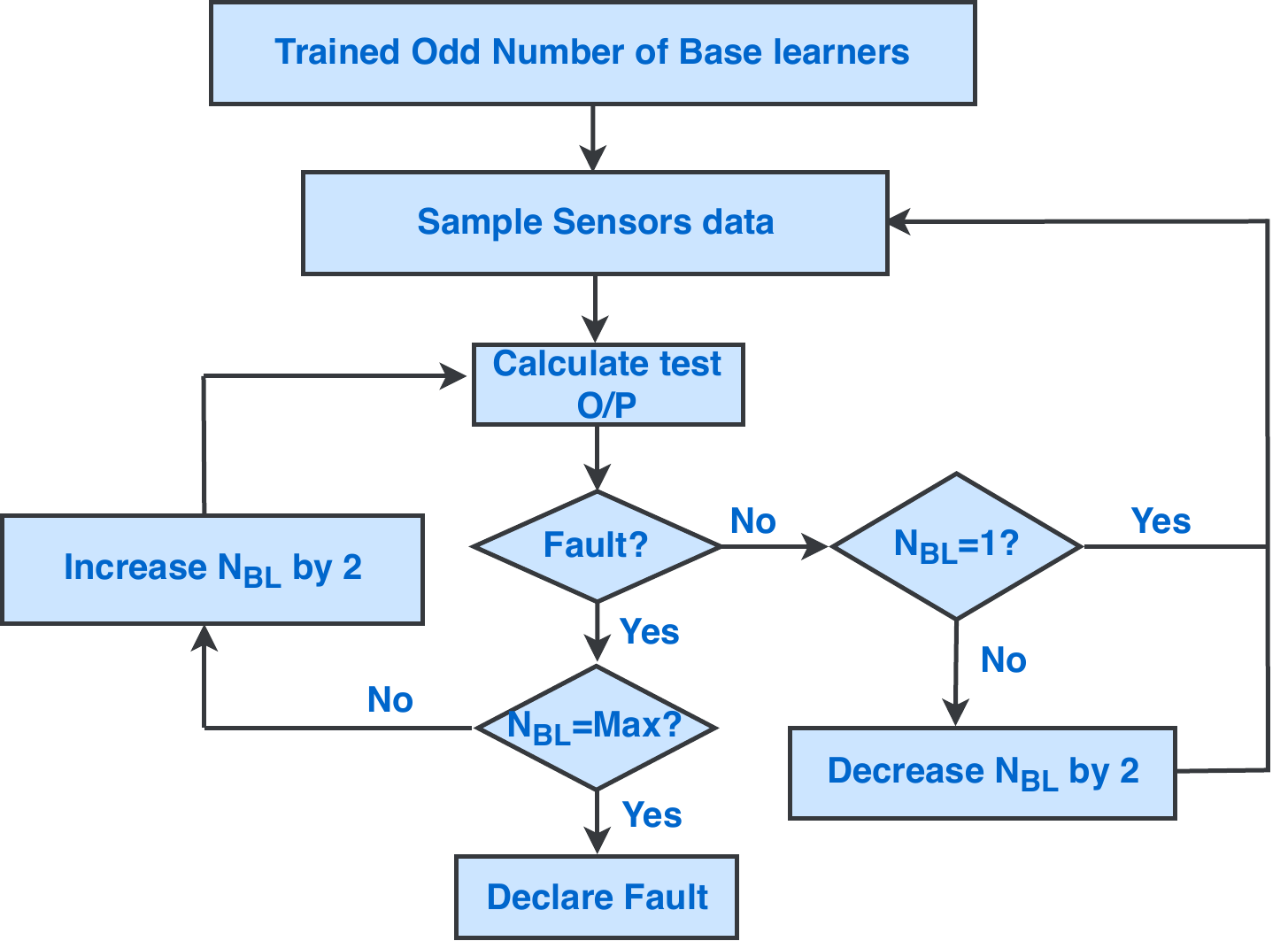}
\caption{Flowchart for ADEPOS: Approximate computing is achieved by adapting the number of hidden neurons in an Ensemble Learning architecture~\cite{Bose_2019}.}
\vspace{-1.5em}
\label{fig_blc}
\end{figure}

The most naive implementation of DNS is to train and keep multiple networks of varying size on the chip with the appropriate network being ON based on the ADEPOS control. While this enables power savings, it is very wasteful of chip area. Hence, it will be best if we can train a bigger network in the learning phase and shut down parts of the network dynamically. However, training a larger ELM network of dimension $L$ and pruning some neurons dynamically lead to erroneous result since during training value of $i^{th}$ neuron weights are dependent on remaining neurons in the network. One of the ways to mitigate this problem is to train $N$ such mini-ELMs each having $L$ hidden neurons and ensemble~\cite{Hansen} outputs using majority voting to get the final output (see figure \ref{fig_ensm}). We call each of these mini-ELM as a base learner (BL). In related work,~\cite{8350948} has applied AdaBoost to increase the resiliency of the overall system but has not explored in the energy reduction perspective. Moreover, the idea of DNS is completely different from neurons and weights pruning~\cite{MoyaRueda2017} performed during the training phase of the network to reduce the number of redundant neurons and parameters. Nevertheless, we can apply such methods to reduce each BL size.

The flowchart in Fig. \ref{fig_blc} shows the steps involved in adaptive network expansion using an ensemble of multiple base learners. At any instance of time, the effective number of neurons in the network is $L_{eff}$, is given by:
\begin{equation} 
\label{eq:Leff}
L_{eff}=L\times N_{BL}
\end{equation}

Where $N_{BL}$ is the number of active base learners in the network. We trained all BLs using healthy data from machine but with different random weights in the first layer. At the beginning of the inference phase, we start with an odd number of BLs. Based on the output of majority voting of all the BLs, we add or remove 2 more BLs from the network since an odd number of BLs are required for majority voting. If at an intermediate step the voting indicates an anomaly, we increase the number of BLs to check whether it is a false alarm due to inaccuracies of approximate computing or a true warning. If all the BLs are deployed and the anomaly detector still raises alarm, we schedule the maintenance. For ASIC implementation, the power supply of all the inactive BLs can be lowered to a point $V_{retention}$ where it can retain its memory content. However, in our present design, all the BLs are implemented in program memory where we modulate the execution time and energy drawn from the battery based on the number of selected BLs.

\subsection{Hidden Neuron Generation using the minimum number of MAC operations}
We define a list of symbols in the following that will be used throughout this subsection:

\begin{tabular}{ll}
$\#$     &Number of\\
$d$     &\# input features\\
$BL$    &Base learner \\
$L$     &\# neurons in each BL \\
$N_{BL,Max}$ & Maximum \# BLs in the network \\

$NG$        &  Neuron generation  \\
$OP^1_{orig}$ &  \# operations in the input layer without NG  \\
$OP^2_{orig}$ &  \# operations in the output layer\\
$OP_{orig}$ &  Total \# operations without NG\\
$OP^1_{NG}$ &  \# operations in the input layer with NG  \\
$OP_{NG}$ &  Total \# operations with NG\\
$L_{phy}$ &  \# physical neurons  \\
\\
\end{tabular}

ADEPOS should be applied over and above other optimization schemes to improve neural network implementation efficiency. Here, we use a NG technique from a recently reported ELM chip~\cite{Chen2019} and show that it can provide benefits in microprocessor-based implementations as well. As mentioned earlier, we randomly choose input layer parameters for ELM computing framework. In general, we require $OP^1_{orig}$~=~$2N_{BL,Max}\cdot d\cdot L$ multiply and addition operations in the input layer (see Eq. (\ref{eqn1a})) and $OP^2_{orig}$~=~$N_{BL,Max} \cdot L$~+~$(L-1) \cdot N_{BL,Max}$ multiply and addition operations (see Eq. (\ref{eqn1c})) in the output layer calculation. 
Here we assume that the system is operating in boundary mode and ignore the calculation related to activation function implementation. Hence,  the total number of multiply and addition operations, $OP_{orig}$, is given by:

\begin{align}
OP_{orig}&= OP^1_{orig}+OP^2_{orig}\notag\\
&= N_{BL,Max}(2dL+ 2L-1)\label{eqn5a}
\end{align}

Following NG scheme, we can generate activation of a virtual neuron by subtracting Eq. (\ref{eqn5b}) and (\ref{eqn5c}) as shown in Eq. (\ref{eqn5d}).

\begin{align}
 h_j&=\sum\limits_{i=1}^dW_{ji}x_i+b_j \label{eqn5b} \\
  h_k&=\sum\limits_{i=1}^dW_{ki}x_i+b_k \label{eqn5c} \\
    h_{jk}&=\sum\limits_{i=1}^d(W_{ki}-W_{ji})x_i+b_k-b_j \label{eqn5d} 
\end{align}
 The operation in Eq. (\ref{eqn5d}) only changes the variance of random weight and bias distribution which does not have any impact on classifier as long as weights and biases are random. The advantage of this method is that by selecting and subtracting any two out of $L_{phy}$ physical neurons, we can create $\frac{L_{phy} \cdot (L_{phy}-1)}{2}$ virtual neurons. In other words, to generate $N_{BL,Max} \cdot L$ virtual neurons, we need approximately $L_{phy}$ physical neurons which is presented in Eq. (\ref{eqn5ddd}). 
 \begin{align}
 L_{phy}&=\ceil*{\frac{1+\sqrt {1+  8\cdot N_{BL,Max} \cdot L } }{2}}\label{eqn5dd} \\
         &\approx\sqrt { 2\cdot N_{BL,Max} \cdot L } \label{eqn5ddd} 
 \end{align}
\begin{figure}[t]
\includegraphics[scale=0.63]{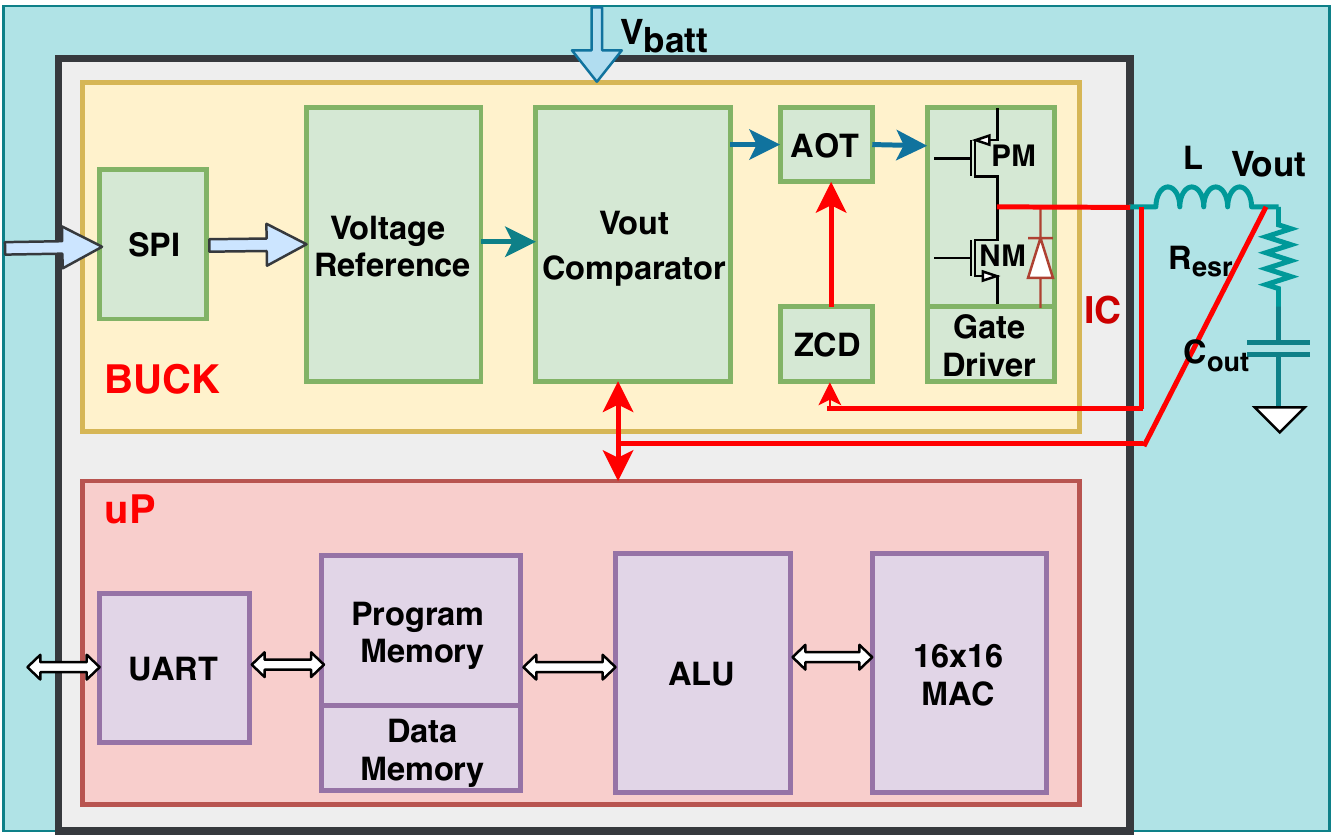}
\caption{ Block diagram of our fabricated integrated circuit which have been used to demonstrate the ADEPOS algorithm. The lower part shows the uP that runs the code for implementing the ELM-B based OCC and ADEPOS controller for DNS, while the upper part depicts the DC-DC converter for DVS.}
\label{top_archetecture}
\end{figure}
At $2\cdot  N_{BL,Max}\leq L$, we can conclude from Eq. (\ref{eqn5ddd}) that $L_{phy}\leq L$. Therefore, the operations required in the first layer, $OP^1_{NG}$, as represented in Eq. (\ref{eqn5ee}) changes drastically. However, the number of calculation in the output layer does not change. Therefore, overall multiply and addition operations, $OP_{NG}$ are needed as shown in Eq. (\ref{eqn5e}) which surely reduces program execution time and burden on multiply and accumulate (MAC) block provided $N_{BL,Max}$ is greater than $1$.

\begin{align}
OP^1_{NG}&= 2L_{phy}\cdot d + N_{BL,Max} \cdot L \label{eqn5ee}\\
OP_{NG}&=OP^1_{NG}+OP^2_{orig}\notag\\
&= 2L_{phy}\cdot d + 3 \cdot N_{BL,Max} \cdot L-N_{BL,Max}\label{eqn5e}
\end{align}

\section{Hardware Architecture}
\label{hw}
\begin{figure}[b]
\centering
\includegraphics[scale=0.14]{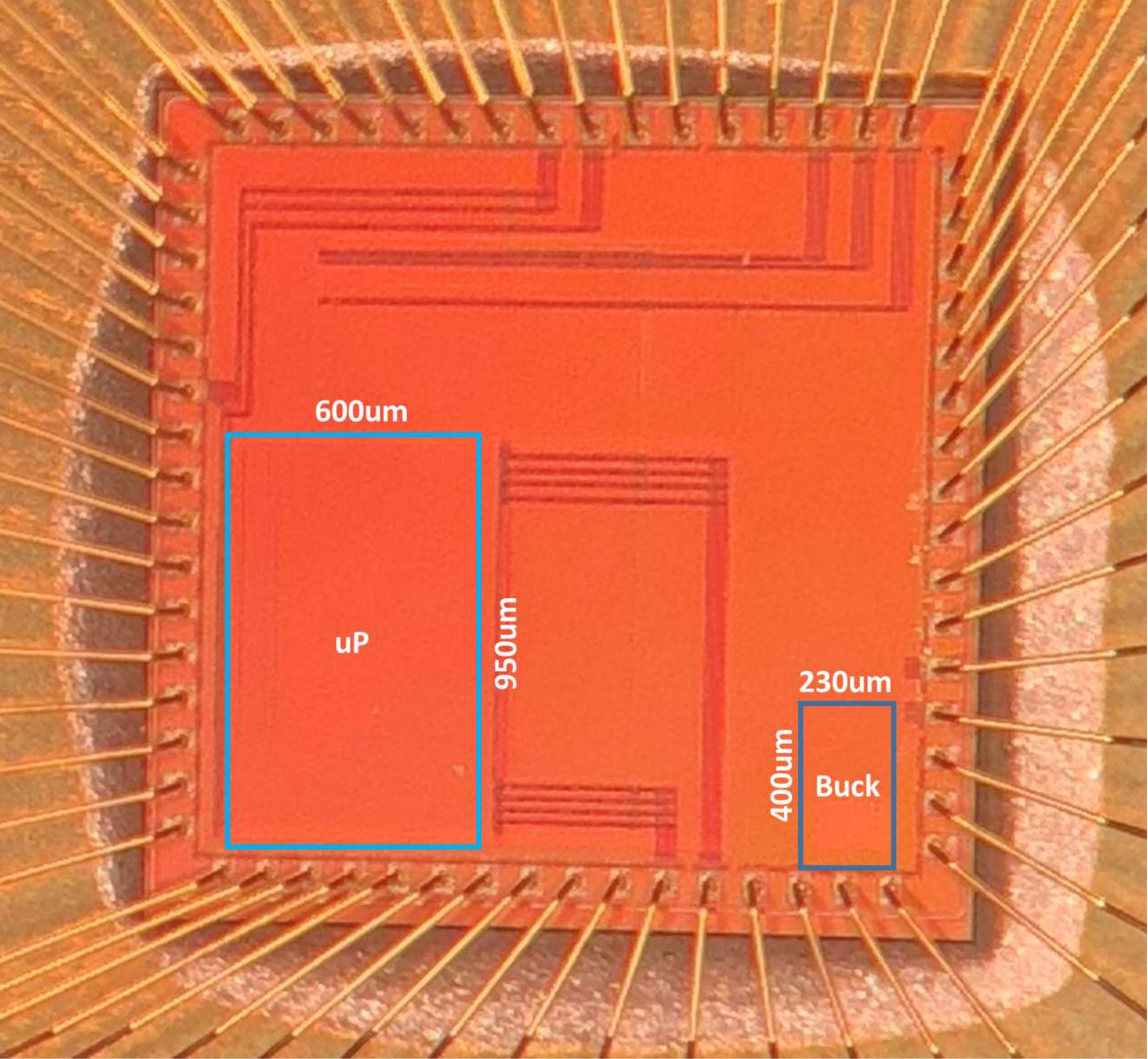}
\caption{ Die photograph of $2$mmX$2$mm chip showing two major blocks- microprocessor(uP) and buck converter.}
\label{die_photo}
\end{figure}

The hardware designed to test ADEPOS consists of two main components: ($1$) a uP following the OpenMSP430 architecture~\cite{openMsp_ref} and ($2$) a DC-DC converter for applying DVS to the uP. Figure \ref{top_archetecture} depicts the overall architecture of the chip while Fig. \ref{die_photo} shows a die photo. The details of these blocks are described next.
\subsection{Microprocessor(uP) Core}
Along with the basic building blocks of a microprocessor, we have incorporated $16$x$16$ multiply and accumulation block (MAC) in the design which is capable of doing $16$bit signed and unsigned MAC operation. The $16$x$16$ MAC block works in parallel with ALU and reduces the executable code size. The program related to all BLs and ADEPOS algorithm are stored in program memory using UART communication protocol and executed serially based on the number of BLs selected out of $N_{BL,Max}$. The uP core has been implemented in UMC $~65$nm CMOS process using core devices.

\subsection{DC-DC Converter}
The DC-DC converter is designed using IO CMOS devices to mitigate the reliability issue due to $3.3$V battery supply. The output of DC-DC converter is configurable from $500$mV to $1250$mV in steps of $50$mV by programming voltage references via SPI communication port. To reduce the overall standby power of the switching converter, we adopted leakage based voltage reference \cite{6293917} and bias current design approach. Once $V_{out}$ voltage falls below the reference voltage, $V_{out}$ comparator turns ON the PMOS switch (PM in Fig. \ref{top_archetecture}) inside the gate driver to charge the output capacitor. Synchronous buck converter topology is followed to minimize power loss in free wheel diode (body diode of NM in Fig. \ref{top_archetecture}). The zero current detector (ZCD) turns OFF the NMOS parallel to free wheel diode whenever the inductor current crosses zero. The ZCD output signal goes to adaptive on time (AOT) block for further processing.

Since the load current requirement of ADEPOS digital core is less than $1$mA, we choose the discontinuous mode of operation (DCM) for the buck converter. Lower supply ripple at the near-threshold region is one of the major design parameters for any system design. On the contrary, buck converter operating at DCM and having constant on time (COT) topology exhibits the maximum output ripple at the maximum battery voltage and the minimum output voltage due to the highest inductor peak current. The output voltage ripple of the buck converter is represented as:


\begin{align}
\Delta V =(R_{esr}+\frac{T_{on}\cdot V_{batt}}{2C_{out} \cdot V_{out}})\cdot \frac{(V_{batt}-V_{out})\cdot T_{on} }{L}\label{eqn5g}
\end{align}

\begin{figure}[b]
\includegraphics[scale=0.7]{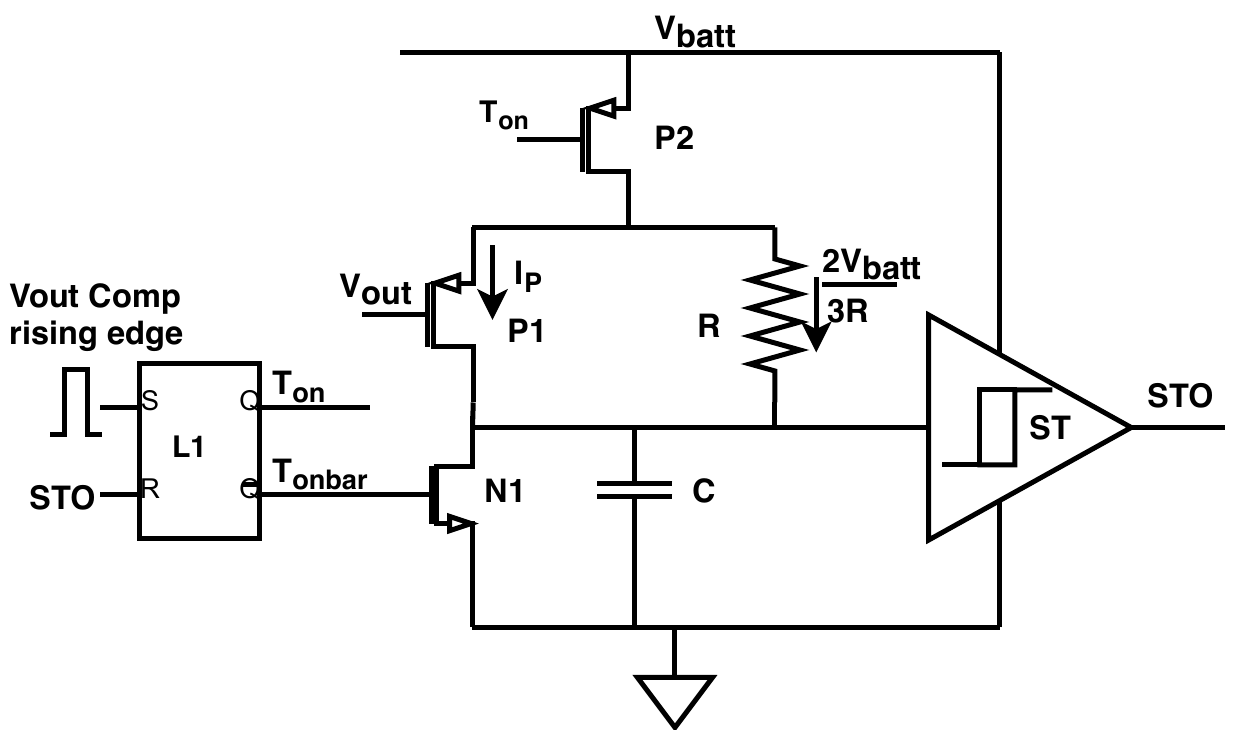}
\caption{ Adaptive on time circuit (AOT) to modulate inductor peak current. Upper and lower threshold of Schmitt trigger are $\frac{2\cdot V_{batt}}{3}$ and $\frac{V_{batt}}{3}$ respectively. Rising edge of the $V_{out}$ comparator output sets the latch and turns ON P2 to enable the charging of C and the rising edge of Schmitt trigger output (STO) resets the latch.}
\label{ripple}
\end{figure}
\begin{figure}[t]
\includegraphics[scale=0.36]{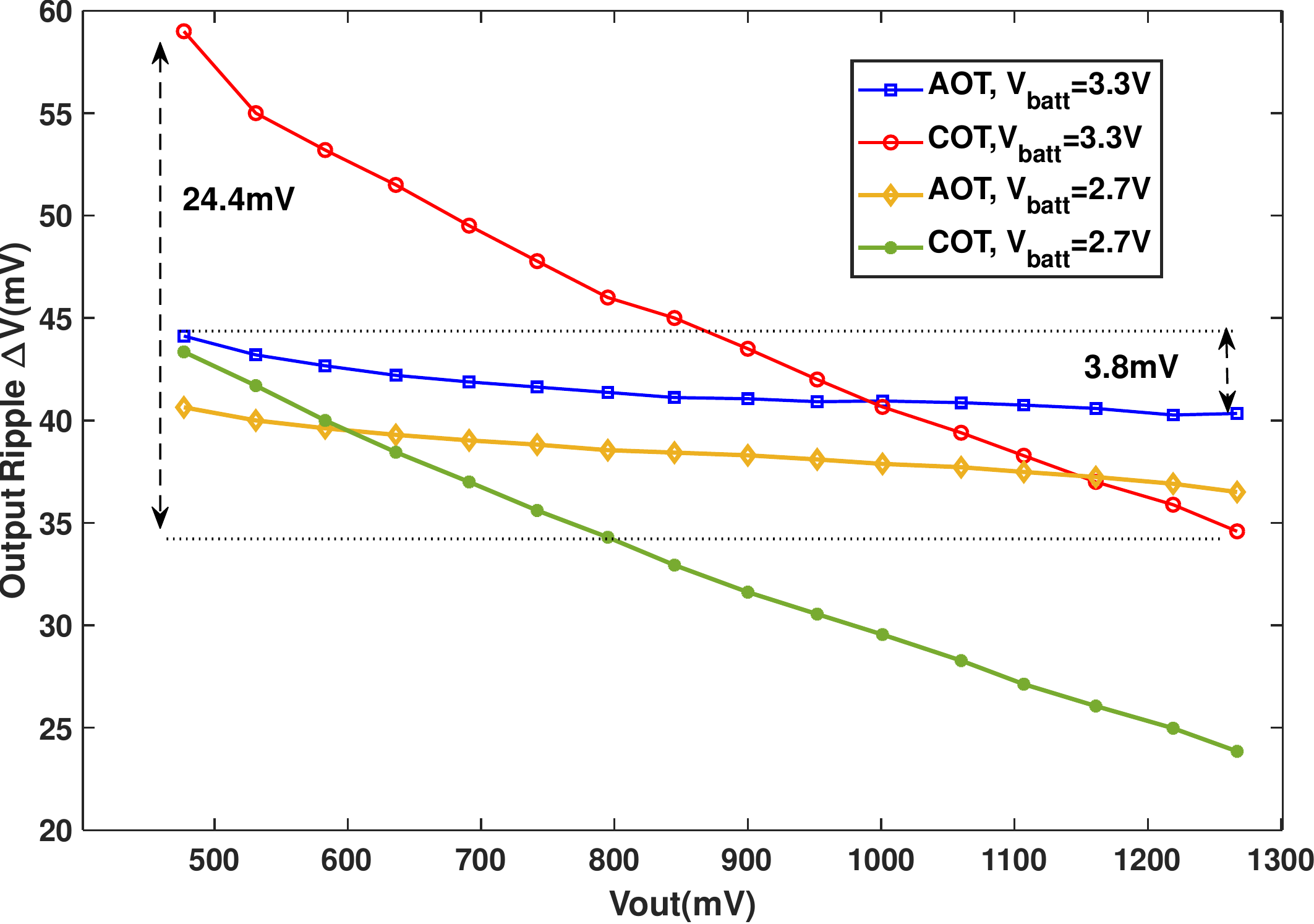}
\caption{ Simulated output voltage ripple of the buck converter with AOT and COT circuit at $R_{esr}$=$100$m$\Omega$, and $C_{out}$=$10\mu$F. Output voltage ripple, $\Delta$V increases drastically when the system operates at lower $V_{out}$ in constant on time mode.}
\label{ripple_output}
\end{figure}

where $R_{esr}$ is the equivalent series resistor of output capacitor $C_{out}$, $T_{on}$ denotes the charging time of $C_{out}$ which is constant for COT topology and $V_{batt}$ and $V_{out}$ are the battery and output voltage of buck converter respectively. Opamp based adaptive on time (AOT)~\cite{Nam2012}, background calibration technique \cite{8351256} have been proposed to take care of the output voltage ripple. Similarly,~\cite{Sahu2007} proposed NMOS and current mirror-based AOT circuit which only takes care of the effect of $V_{batt}$ variation on the output ripple. But a battery-operated system which employs dynamic output voltage scaling should mitigate the effect of both $V_{batt}$ and $V_{out}$ variation on the output ripple. In this design, we suggested single PMOS transistor based adaptive on time  (see Fig. \ref{ripple}), $T_{on}$, which follows Eq. (\ref{eqn5h}) and keeps the output ripple almost constant as highlighted in Fig.\ref{ripple_output} and consumes very low power.
\begin{align}
T_{on} = \frac{k1 \cdot V_{batt}}{(k2 \cdot V_{batt}- k3 \cdot V_{out})+k4} \label{eqn5h}
\end{align}
Parameters $k1$, $k2$, $k3$ and $k4$ are  dependent on process, transistor aspect ratio, resistor, R, and capacitor, C. We discuss the derivation of  Eq. (\ref{eqn5h}) and its related parameter in appendix.

The current through P1 and R charges the capacitor C. Whenever the voltage across the capacitor C crosses the upper threshold voltage ($\frac{2\cdot V_{batt}}{3}$) of the  Schmitt trigger (ST), its output voltage goes high and resets the latch. The process variation of PMOS can be taken care by varying the aspect ratio of P1 and resistor R.

The circuit is simulated in SPICE using UMC $65$nm models. In Fig. \ref{ripple_output}, we have plotted output voltage ripple for AOT and COT (constant on time) approaches at $R_{esr}$=$100$m$\Omega$, $C_{out}$= $10\mu$F, $L$=$2.2\mu$H and $V_{batt}=3.3$V and $2.7$V across different $V_{out}$ voltages. It can be seen that the output ripple variation is $24.4$mV across $V_{out}$ and even twice if we consider $V_{batt}$ variation in COT approach. On the contrary, the proposed AOT circuit keeps the output ripple almost constant across $V_{batt}$ and $V_{out}$ and draws only $50$nA dynamic current when the load is $1$mA.
\section{Results}
\label{re}

\subsection{Dataset}
\label{data}
In order to validate the proposed ADEPOS algorithm, we use NASA bearing dataset \cite{NasaDataset} provided by the Center for Intelligent Maintenance Systems (IMS), University of Cincinnati. 
The bearing vibration data was sampled at $20$KHz for $1$s duration in an interval of $10$ minutes. Out of twelve bearings, two bearings faced outer race failure and two of them suffered inner race and roller element defect respectively.
The summary of time series data of bearings is captured in Table \ref{tab2}.

\begin{table}[t]
\centering
\caption{Summary of NASA bearing time series dataset}
\resizebox{\columnwidth}{!}{
    \begin{tabular} {|M{.9cm}|M{.9cm}|M{.8cm}|M{1.4cm}|M{.9cm}|}\hline
        Test &Bearings &\# Files &\# Data point in each file &Remarks\\ \hline
        \multirow {4}{*} {Dataset1} &Bearing1 &2156 &20480 & Healthy \\
        \cline{2-5}
              &Bearing2 &2156 &20480 & Healthy \\
        \cline{2-5}
            &Bearing3 &2156 &20480 & Faulty \\
        \cline{2-5}
             &Bearing4 &2156 &20480 & Faulty \\ \hline
        \multirow {4}{*} {Dataset2} &Bearing1 &984 &20480 & Faulty \\
        \cline{2-5}
              &Bearing2 &984 &20480 & Healthy \\
        \cline{2-5}
            &Bearing3 &984 &20480 & Healthy \\
        \cline{2-5}
             &Bearing4 &984 &20480 & Healthy \\ \hline
        \multirow {4}{*} {Dataset3} &Bearing1 &6324 &20480 & Healthy \\
        \cline{2-5}
              &Bearing2 &6324 &20480 & Healthy \\
        \cline{2-5}
            &Bearing3 &6324 &20480 & Faulty \\
        \cline{2-5}
             &Bearing4 &6324 &20480 & Healthy \\ \hline

       \end{tabular}
    }
\label{tab2}
\end{table}
 Statistical features such as mean, root mean square (RMS), kurtosis~\cite{MARTIN199567} from the time domain, FFT of fixed sub-band frequencies~\cite{873206} from the frequency domain and wavelet packet transformation (DWT)~\cite{PENG2004199} from the time\textendash frequency domain are popular in the literature as intermediate data for machine health monitoring. In this work, we conduct our experiments on five time domain features extracted from the raw vibration data contained in NASA dataset, such as (a) RMS, (b) Kurtosis, (c) Peak-Peak, (d) Crest factor, and (e) Skewness since we have validated that this is a minimal set of features that are the most informative. Moreover, the time complexity of calculating $m$ such features is $m\cdot\mathcal{O}(n)$ where $n$ denotes the number of data points in each file.
 \begin{table*}[h!]
\centering
\caption{Convergence study of different OCC methods}
    \begin{tabular} {|M{3cm}|M{1.8cm}|M{1.8cm}|M{1.8cm}|M{1.8cm}|M{2cm}|M{2cm}|}\hline
    &\multicolumn{4}{|c|}{ ELM} &\multicolumn{2}{|c|}{ Traditional AE(TAE)}\\ \hline
          &OSELM-B  &OSELM-AE &OPIUM-B &OPIUM-AE &TAE-Un & TAE-Ti \\ \hline
        Operation(Million) &0.59	&1.08 &0.26 &0.61 &0.97 &0.24 \\ \hline
        Data Memory 	&1262 &1350 &482  &570 &310 & 310\\ \hline
        Robustness Margin($\rho$) 	&94 &127 &56  &163 &20 & 1.9\\ \hline
    \end{tabular}
\label{tab1x}
\end{table*}
\subsection{A comparative study of ELM-based algorithms and traditional AEs}

\label{sec:AE_compare}
A comparative convergence study of online training among ELM and traditional methods in bearings~\cite{NasaDataset} health monitoring is summarized in Table \ref{tab1x}. In this analysis, we train below mentioned ML models in MATLAB using a few hundred samples of bearing data and when the convergence of $\beta$ is reached, we use the remaining samples for calculating testing error ($\epsilon$). We use boundary and reconstruction OCC framework for OSELM and OPIUM based online learning approach and are termed as OSELM-B, OSELM-AE, OPIUM-B, and OPIUM-AE respectively. Since OSELM method demands matrix inverse operation, its computational complexity ($\mathcal{O}(L^3)$) is higher than OPIUM ($\mathcal{O}(L^2)$). Hence, OSELM-B needs $2.27$X number of operations (operations/sample X Number of samples) and $2.6$X higher SRAM memory to converge using the LU decomposition method than its counterpart (OPIUM-B). In terms of the number of operations and SRAM memory requirement, the performance of OPIUM-B is comparable with both variants of traditional autoencoder - (a) traditional untied autoencoder (TAE-Un) \cite{tae_untied} and (b) tied autoencoder (TAE-Ti) \cite{tae_tied}. We use ADAM~\cite{Kingma} optimizer to train the TAEs. 

In order to further compare all the algorithms, we define robustness margin ($\rho$) as shown in Fig. \ref{robust} that indicates the separation margin between healthy and faulty bearings and the feasibility of placing threshold to completely separate the faulty bearings from a healthy one. Parameter $\gamma$ in Fig. \ref{robust} is defined by:
\begin{figure}[t]
\includegraphics[scale=0.32]{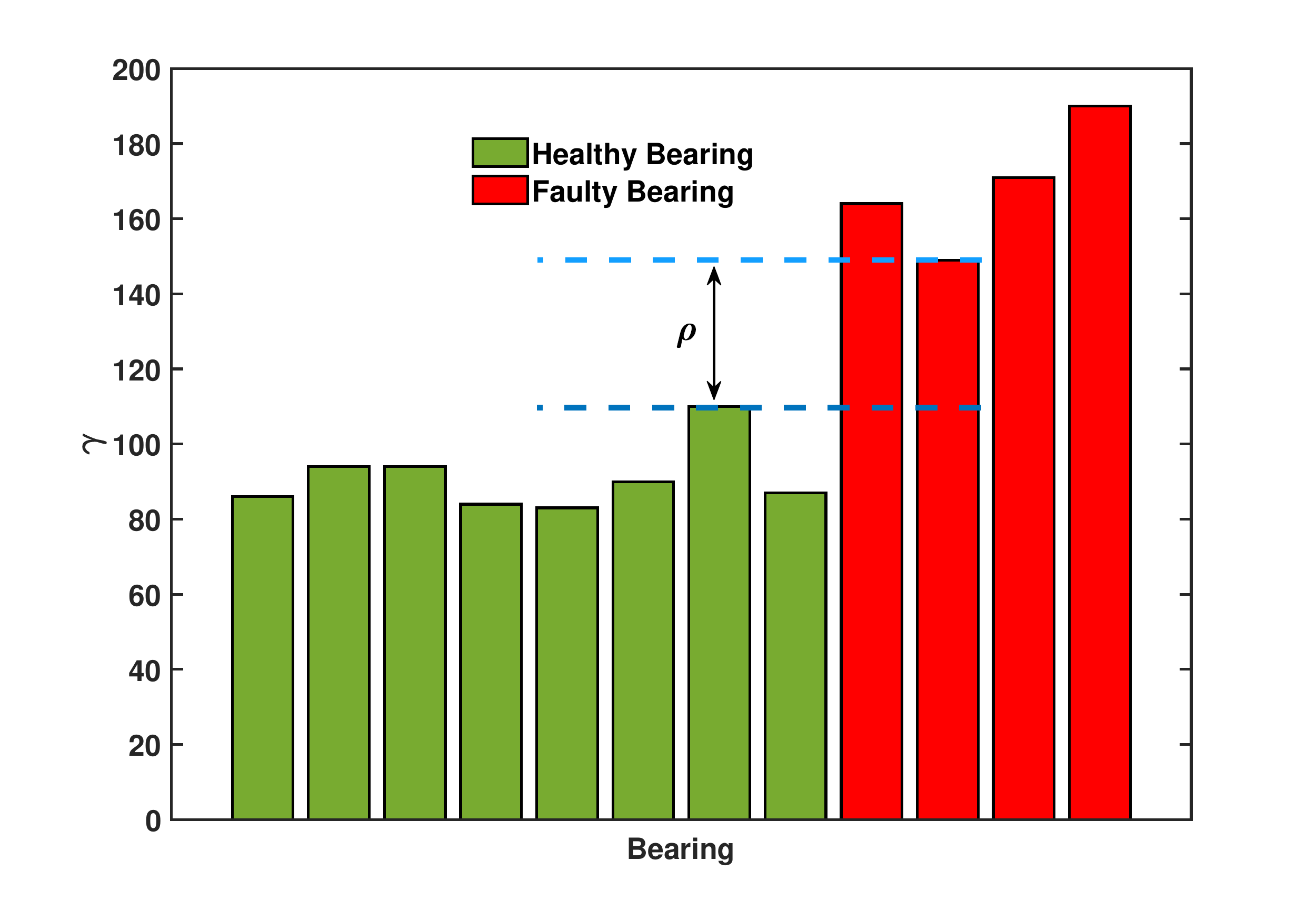}
\caption{Robustness ($\rho$) margin indicates the separation margin between healthy and faulty bearings and the feasibility of placing threshold to completely separate the faulty bearings from a healthy one. Parameter $\gamma$ is defined in Eq. (\ref{eqn6ax}).}
\label{robust}
\end{figure}
\begin{equation}
 \begin{split}
\gamma_i=\frac{{\vert \epsilon^{test}\vert}_{max}}{<\vert\epsilon^{train}\vert>}; ~~i=1,2,...,12 
\label{eqn6ax}
 \end{split}
\end{equation}
where $<\vert\epsilon^{train}\vert>$ is the indication of average noise in the data during healthy operation and $\gamma$ indicates how much deviation can happen due to degradation of condition. Ideally, the healthy bearings should have low values of $\gamma$.

In terms of robustness margin ($\rho$) OPIUM-B shows better performance than both TAEs. Hence, we choose OPIUM-B algorithm as OCC for machine health monitoring.
\subsection{Threshold Selection}
Due to the limitation of unhealthy bearing data and time stamp of failure, we follow the leave-one-out (LOO) approach to calculate the threshold value ($\lambda$) using Eq. (\ref{eqn6a}). We divide $12$ bearings into two groups, $11$ bearings for threshold calculation and the remaining bearing for testing deploying the calculated threshold ($\lambda$). Errors ($\epsilon$) (difference of expected value and output of boundary-OCC) from good bearings out of all $11$ training bearings data are used for threshold calculation. We carry out the LOO strategy for each of the $12$ bearings keeping one bearing data for testing.
\begin{equation}
\lambda=Max(\epsilon)+0.5\times c\times \sigma_{\epsilon}
\label{eqn6a}
\end{equation}
$Max(\epsilon)$ and $\sigma_{\epsilon}$ are the maximum and standard deviation of $\epsilon$ values of good bearings in the training set. We choose $c=1$, since at this value AUC (area under the curve) of ROC curve is one. 

\subsection{Optimal Neurons ($L$) and $N_{BL,Max}$ Selection}\label{lselection}
\begin{figure}[t]
\includegraphics[scale=0.34]{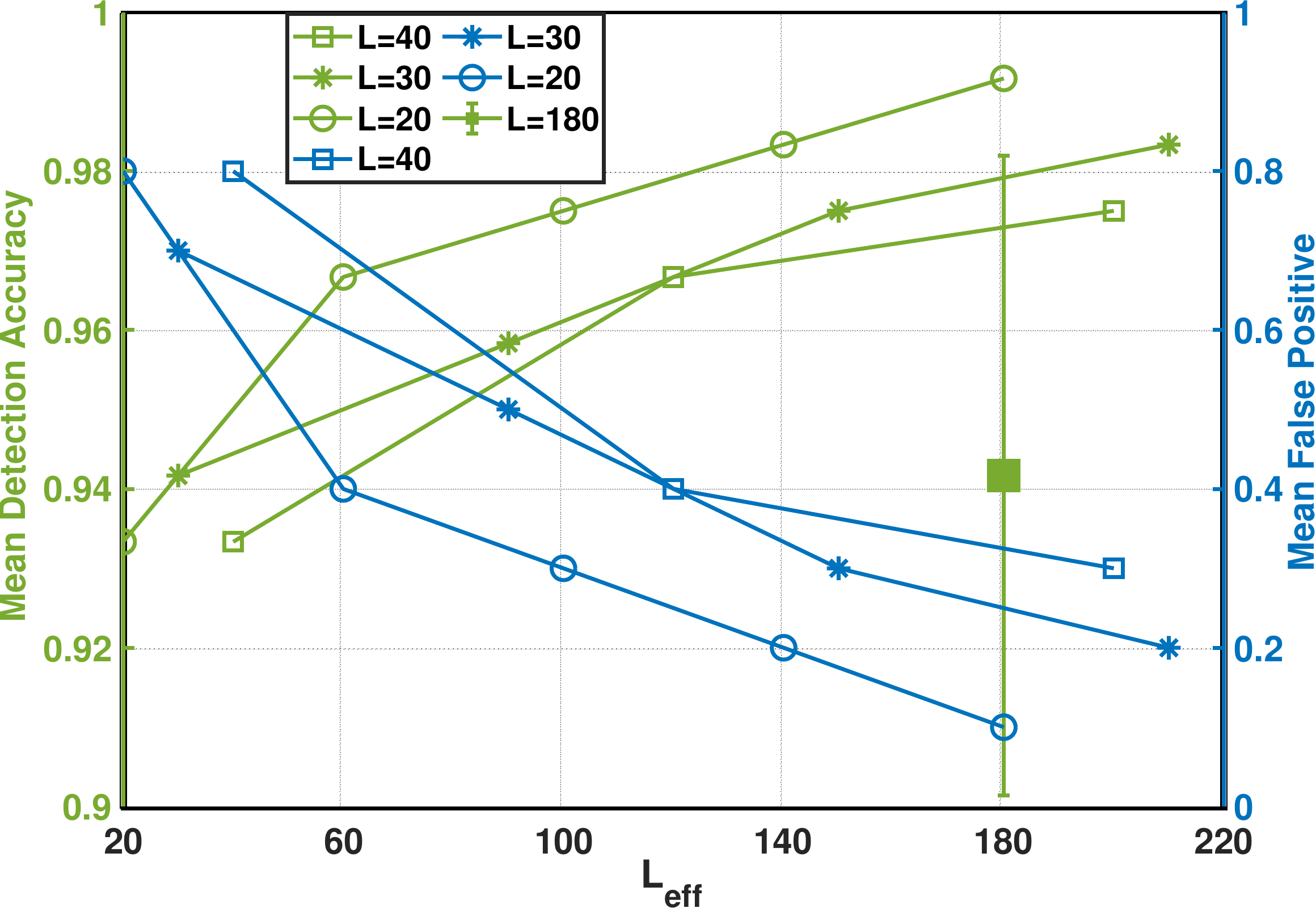}
\caption{Mean detection accuracy and mean false positive vs. effective number of hidden neurons ($L_{eff}$) in the network. Detection accuracy increases with $L_{eff}$ and smaller value of $L$. }
\label{accuracy}
\end{figure}
It is shown in~\cite{Bose_2019} that at $L=20$ and  $N_{BL,Max}=9$, the mean detection accuracy is almost $100\%$ over $10$ trials (for $10$ different random seeds). The experimental results are also plotted in Fig. \ref{accuracy} for different $L=(20,30,40)$ and $N_{BL,Max}=(9,7,5)$ values using ensemble of BLs. It is apparent from Fig. \ref{accuracy} that mean detection accuracy is monotonically increasing with an effective number of hidden neurons $L_{eff}$ in the network and smaller values of $L$.  Furthermore, the mean false positive over $10$ trials decreases if higher number of BLs are ensemble together. The vertical green line at $L=180$ shows the detection accuracy (mean$\pm$SD) of a single large classifier. It can be seen that the mean detection accuracy of a single large classifier having L=$180$ neurons is lower than that of an ensemble of BLs with a smaller number of neurons. Hence, an ensemble of BLs not only increases the detection accuracy but also provides an opportunity to increase the energy saving of the system by leveraging the proposed ADEPOS algorithm. Since at $L=20$ and  $N_{BL,Max}=9$, mean detection accuracy is almost $100\%$, we choose this configuration for showing the energy saving of our ADEPOS algorithm in uP. Before going into the details of energy saving of the proposed ADEPOS algorithm, we will discuss the characterization of uP and DC-DC converter separately.

\begin{figure}[t]
\includegraphics[scale=0.5]{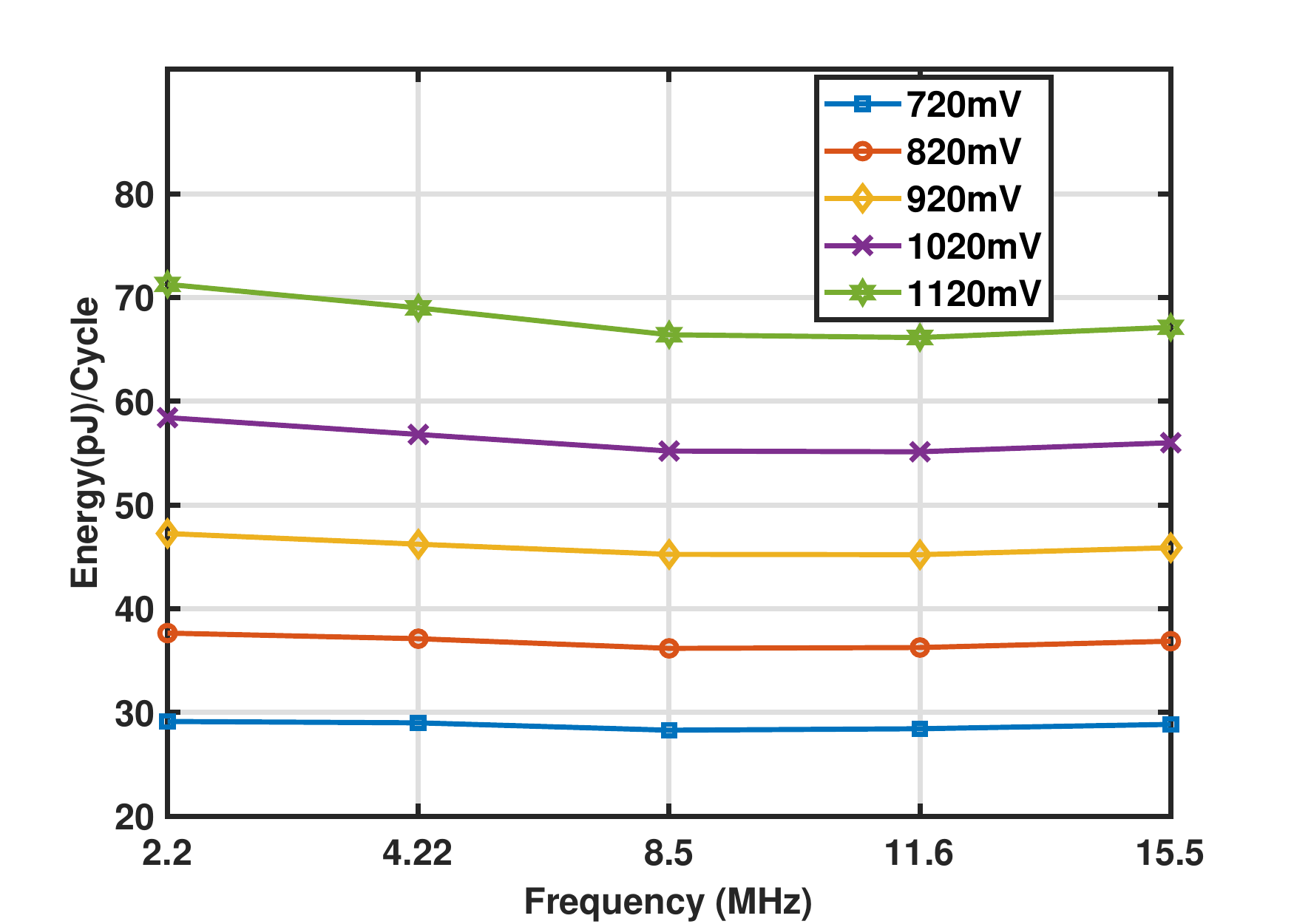}
\caption{Energy per clock cycle of uP across different operating frequencies and voltages.}
\label{char_up}
\end{figure}
\subsection{Chip Characterization}
In Fig. \ref{char_up}, we show the energy per clock cycle of uP at different operating voltages and frequencies. As expected, energy scales with supply voltages but remains almost constant with frequencies. The minimum energy operating point occurs at $720$mV and $8.5$MHz. Although the memory can retain its value till $190$mV, uP core generates fallacious output below $700$mV at higher operating frequencies due to SRAM read or write failure.

\begin{figure}[t]
\includegraphics[scale=0.48]{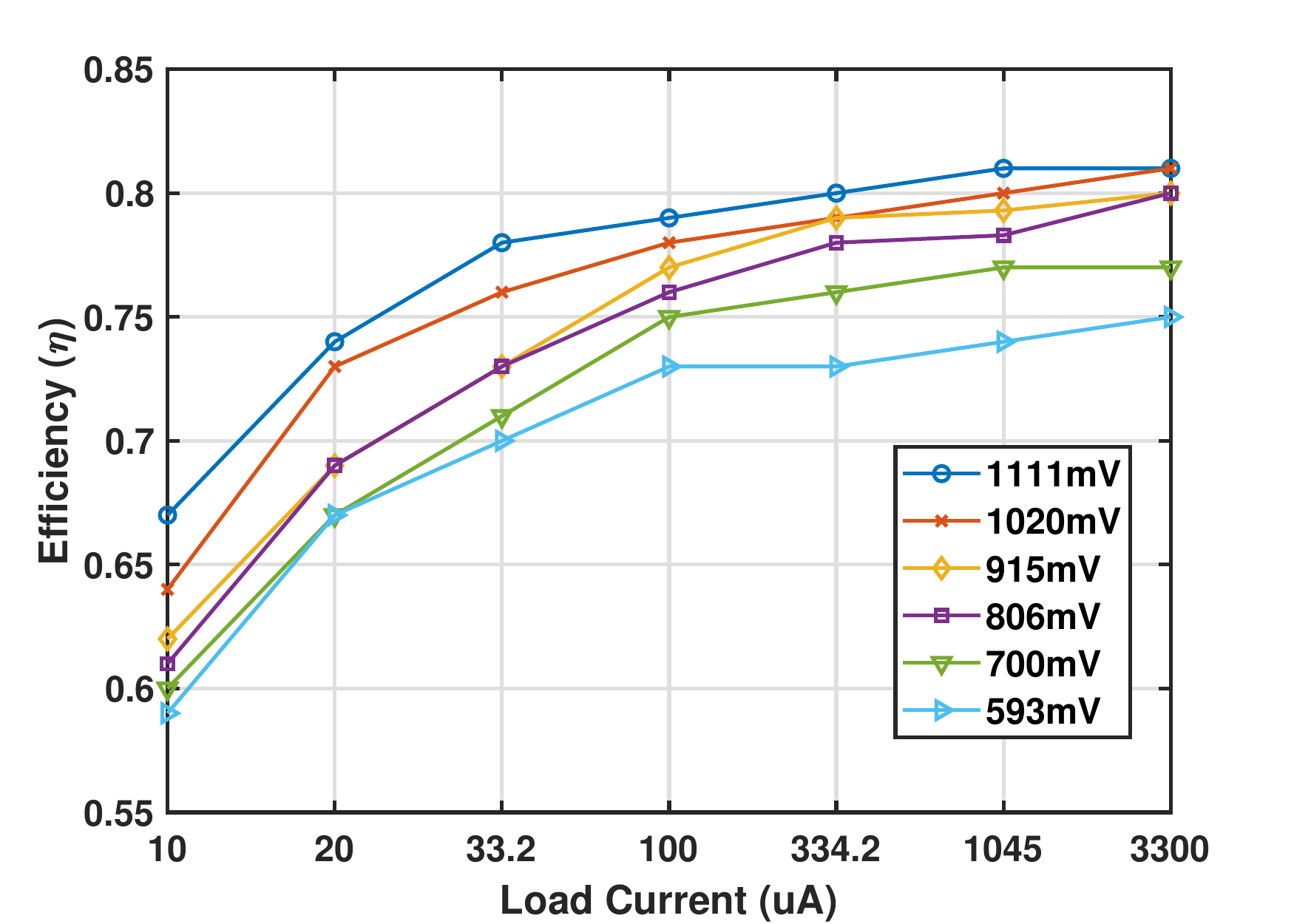}
\caption{Efficiency of DC-DC converter across load current at different output voltages.}
\label{efficiency_curve}
\end{figure}

The Fig. \ref{efficiency_curve} indicates that the efficiency of the DC-DC converter increases with load current. Since there are hardly any changes in energy per cycle between $8.5$MHz and $15.5$MHz and dynamic current of uP increases at higher frequencies, it will be beneficial to operate the system at $15.5$MHz. The efficiency of the buck converter is lower than expected due to high inductor peak current and higher metal resistance from external pads to the NMOS and PMOS transistor of the gate driver. Since the DC-DC converter supports more than $10$mA load current and the required maximum processor current is $1$mA, we can increase the inductor from $2.2\mu$H to $10\mu$H to reduce the peak inductor current which in turns reduces conduction loss across NMOS and PMOS switch of the gate driver.
 
 The overall energy consumption by the predictive maintenance system can be written as:
  \begin{equation}
 \begin{split}
Energy= &\frac{V_{out,sta} \cdot I_{core,sta} \cdot  T_{sleep}}{\eta_{sta}} +\\ & \frac{V_{out,dyn} \cdot I_{core,dyn} \cdot T_{active}}{\eta_{dyn}}
\label{eqn6b}
\end{split}
\end{equation}
where $V_{out,sta}$ and $I_{core,sta}$ are the output voltage of DC-DC converter and static current of processor respectively when the system is in sleep and $V_{out,dyn}$ and $I_{core,dyn}$ are the output voltage of DC-DC converter and static current of processor respectively when the system is active. $\eta_{sta}$ and $\eta_{dyn}$ are the efficiency of the DC-DC converter at two different operating conditions and function of the output voltage and load current. The static and dynamic power of the DC-DC converter is taken into account during the efficiency calculation.

\begin{table*}[h]
\centering
\caption{Performance Comparison of Buck Converter with published Works}
    \begin{tabular} {|M{2.8cm}|M{2.8cm}|M{3cm}|M{2.8cm}|M{2.8cm}|}\hline
         &\cite{TPS62736} &\cite{Fu2016} &\cite{Paidimarri2017} &This Work \\ \hline
         Control Method  &Hysteresis \& PFM  &Hysteresis \& AMOT  &PFM &PFM \& AOT \\ \hline
        Input Voltage($V_{in}$	&2-5.5V &1.8-4.2V &1.2-3.3V &2.3-3.5V\\ \hline
        Output Voltage($V_{o}$) &1.3-5V &0.9-1.4V &0.7-0.9V &0.47-1.25V\\ \hline
       Max Load Current &50mA  &40mA &1mA &10mA\\ \hline
        Output Ripple &20mV &30mV &20mV &41mV \\ \hline
       Quiescent Current &380nA &12uA &\~100pA &700nA \\ \hline
        Inductor (L) &10$\mu$H &4.7$\mu$H &47$\mu$H &2.2$\mu$H \\ \hline
        \multirow {2}{*} {Efficiency at $I_L=1$mA} &$V_{in}$=$3.3$V, $V_{o}$=$1.3$V  &$V_{in}$=$3.3$V, $V_{o}$=$1.2$V & $V_{in}$=$3$V, $V_{o}$=$0.8$V  &$V_{in}$=$3.3$V, $V_{o}$=$0.8$V \\
        \cline{2-5} &87\% &83\% &80\% &78.3\% \\ \hline
    \end{tabular}
\label{tab_per}
\end{table*}
Although the system can operate down to $720$mV, we choose the optimal operating voltage of the system as $750$mV to have some margin. At $750mV$, we will lose around $5\%$ energy due to lower efficiency of DC-DC converter compared to $1111$mV which is apparent from Fig. \ref{efficiency_curve}. However, voltage scaling can save around $53\%$ energy at $15.5MHz$ due to the lower voltage and dynamic current (see Fig. \ref{char_up}) leading to better system level efficiency.

The performance  of the designed buck converter is comparable with recently published work as presented in Table \ref{tab_per}. In general, parasitic losses are lower at the higher output voltage, $V_o$, lower input voltage, $V_{in}$, and higher value of inductor due to the lower inductor peak current at this operating condition. Hence, the efficiency is better for \cite{TPS62736} and \cite{Fu2016}. Since we are using a lower inductor value of $L$=$2.2\mu$H, there is room to improve efficiency using higher inductor value at the cost of lower load current. Lower output ripple is another benefit of using higher inductance.
\subsection{Experimental setup for Validation of ADEPOS algorithm} \label{setup_exp}
\begin{figure}[b!]
\includegraphics[scale=0.73]{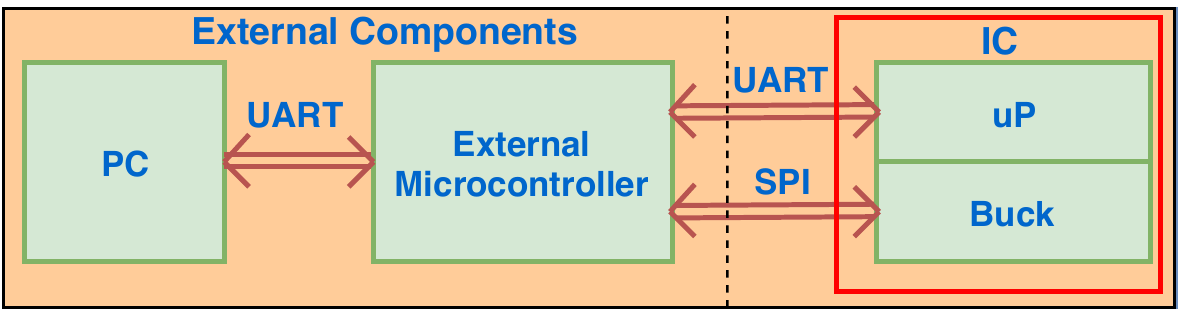}
\caption{Experimental setup for validating ADEPOS algorithm. A PC is used to visualize the output of base learners and send data using Matlab. The external controller acts as an interface between uP and PC and buck converter.} 
\label{setup}
\end{figure}

 In Fig. \ref{setup}, we present the experimental setup for validating ADEPOS. A PC is used to visualize the output of base learners and send data using Matlab. The external controller acts as an interface between uP and other modules (PC and buck converter). It handles data transfer to and from the on-chip uP and also sets the DC-DC converter to $750$mV during operation, and $600$mV during inactive periods. Since the SRAM can retain its value until $190mV$, we can power the uP following HYPNOS technique~\cite{Jayakumar2014} and bring down the power consumption further in the nW range.
\begin{figure*}[t]
\includegraphics[scale=0.5]{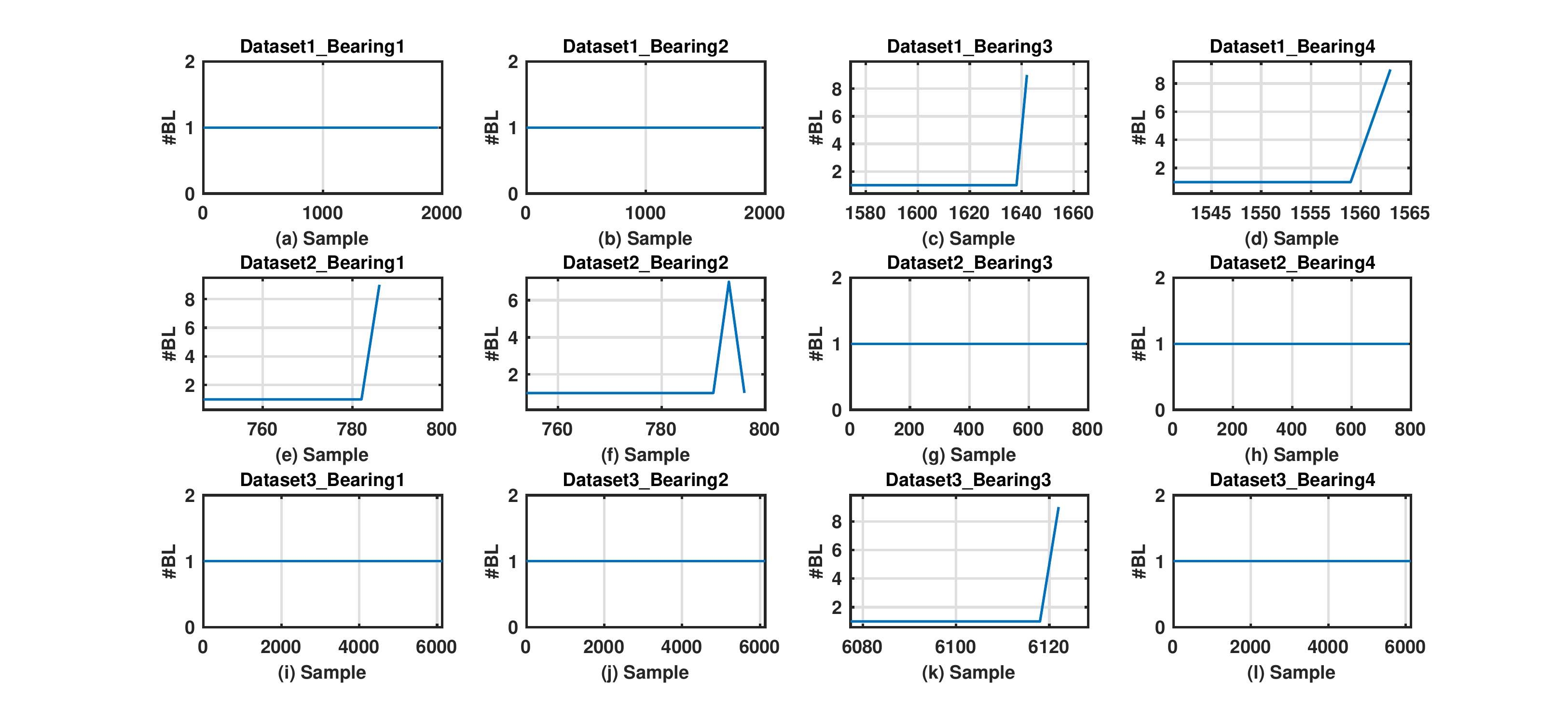}
\caption{Number of active base learners along the lifetime of all 12 bearings time series data for seed value $=10$. Bearings in (c), (d), (e) and (k) failed at the end of their lifetime which is indicated by utilization of all base learners $(9)$ in the system. Healthy bearings in (a), (b), (g), (h), (i), (j), and (l) deploys single base learner except bearing in (f) uses $7$ base learners on sample data to get the correct decision.}
\label{plot_10}
\end{figure*}

Since the uP performs integer operations, the extracted feature from raw bearing data is converted to a 6bit integer value. The first layer random weight, bias and learned second layer weights for $9$ base learners (selection of $L=20$ and  $N_{BL,Max}=9$ have been discussed in section \ref{lselection}) are stored in the program memory of uP core. Once the data transfer is completed between the external controller and data memory, the external controller enables the uP, and when the execution is over, the uP goes back to sleep again. We generate random weight and bias for $10$ different seeds to validate the proposed ADEPOS algorithm on NASA bearings dataset.


Figure \ref{plot_10} shows the number of active base learners along the lifetime of all 12 bearings time series data. Bearings in (c), (d), (e) and (k) failed at the end of their lifetime which is indicated by utilization of all base learners ($9$) available in the system. Healthy bearings in (a), (b), (g), (h), (i), (j), and (l) deploy single base learner along the age of the bearings except bearing in Fig. \ref{plot_10}(f) which deploys $7$ base learners on sample data to get the correct decision.

Based on the adaptive usage of the number of ensembles, we find that the average value of $L_{eff}$ (over $10$ trials) throughout the lifetime of all the bearings is only $20.11$ without sacrificing the accuracy obtained by using $L_{eff}=180$ for $N_{BL}=9$ networks in the ensemble. Thus compared to the case of using a fixed value of $L=180$ neurons, ADEPOS enables $8.95$X reduction in an effective number of neurons. This reduction in neurons translates to the energy reduction in computation as shown next.

\subsection{Energy Measurement}
\begin{figure}[t]
\centering
\includegraphics[scale=0.32]{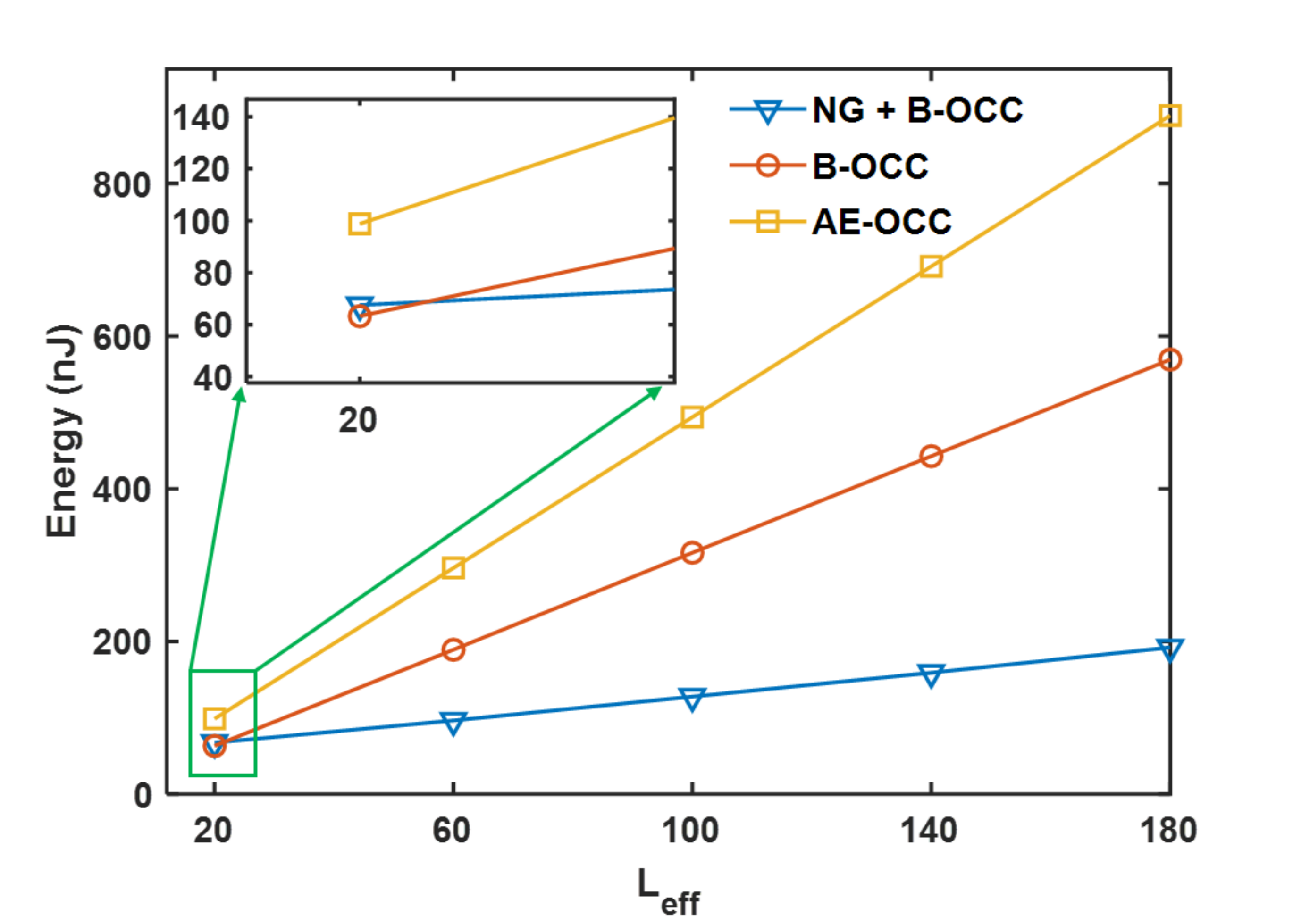}
\caption{Impact of Hidden neurons ($L_{eff}$) on energy consumption for different algorithms. As expected energy consumption increases linearly with hidden neurons in the network but the slope depends on the algorithm.}
\label{energy_bl}
\end{figure}

We use INA$210$~\cite{INA210} IC from Texas instruments to measure the exact execution time of each base learner. In order to calculate the energy consumed by the algorithms(AE-OCC, B-OCC, B-OCC+NG), we measure the execution time of the algorithms employing INA$210$ and the power drawn by the uP to execute these algorithms. Subsequently, we multiply the power and the execution time to get the final energy number. Fig. \ref{energy_bl} shows the impact of hidden neurons ($L_{eff}$) on energy consumption for different algorithms at $750$mV and $15.5$MHz frequency. In system level, these number will scale up by the same factor if we take into account the efficiency of the DC-DC converter at that operating condition. As expected, energy consumption increases linearly with hidden neurons in the network but the slope depends on the algorithm. Energy consumption of the boundary-based OCC(B-OCC) is lower than autoencoder-based OCC (AE-OCC) since B-OCC requires less number of computation at the output layer than AE-OCC does. In neuron generation(NG) approach, replacement of multiplication operation by subtraction operation results in a reduction of execution time and a number of operation of each base learner. The combining effect of execution time and a number of operation reduction results in $~3$X energy reduction for $L_{eff}=180$ in B-OCC. Since, in neuron generation method, we need multiplication along with subtraction for the first base learner, $6.7\%$ more energy is required for the single base learner (L=$20$) which is magnified and shown in the inset.

\begin{figure}[t]
\includegraphics[scale=0.36]{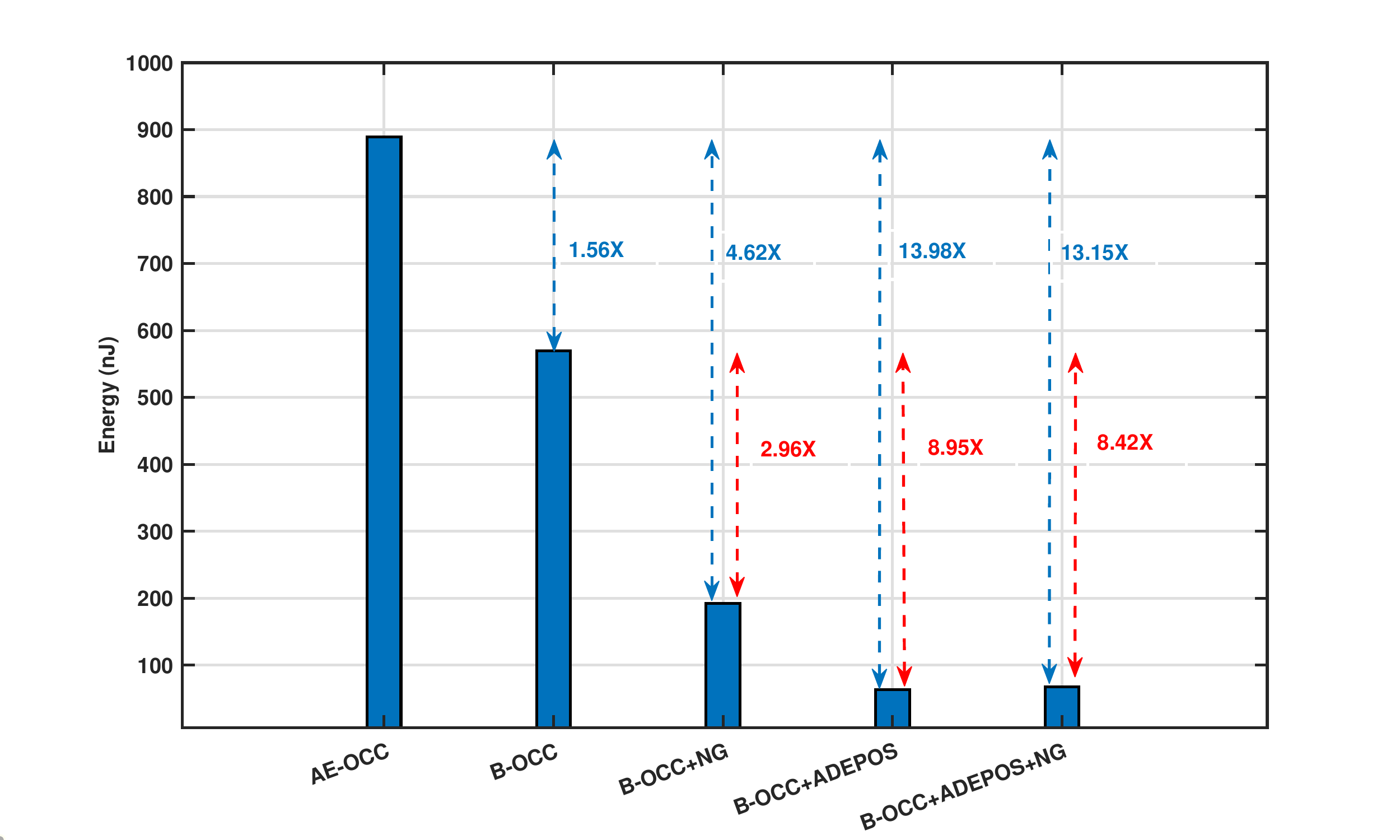}
\caption{Energy saving by use of appropriate algorithm and ADEPOS on NASA bearing time series data. ADEPOS with B-OCC enables $8.95X$ saving in energy.}
\label{energy_bl1}
\end{figure}

Figure \ref{energy_bl1} shows the energy saving by use of the appropriate algorithm and ADEPOS on NASA bearing time series data. B-OCC and B-OCC along with NG enable $1.56$X and $4.62$X energy reduction  respectively compare to AE-OCC for $L_{eff}=180$. Moreover, combining ADEPOS algorithm with B-OCC helps us to achieve $13.98$X less energy consumption than AE-OCC. This is due to a combination of the reduction in number of neurons (Section \ref{setup_exp}) along with reduced number of operations due to the choice of algorithm (B-OCC as opposed to AE-OCC).  
Energy saving of ADEPOS along with NG is slightly less due to two reasons a) L is small ($20$ here for each BL). It can be shown that the energy consumption of single BL at L$> 24$ generated using NG method is lower than that without using NG approach. b) The average number of neurons for NASA bearing dataset is $20.11$ whereas the break-even point is $22.22$ neurons. Even though the energy benefit of neurons generation is not visible here, it requires $64\%$ less flash memory to implement $9$ base learners than B-OCC, since, in B-OCC method, random weights and biases are used as an immediate operand. 
\begin{figure}[b]
\centering
\includegraphics[scale=0.32]{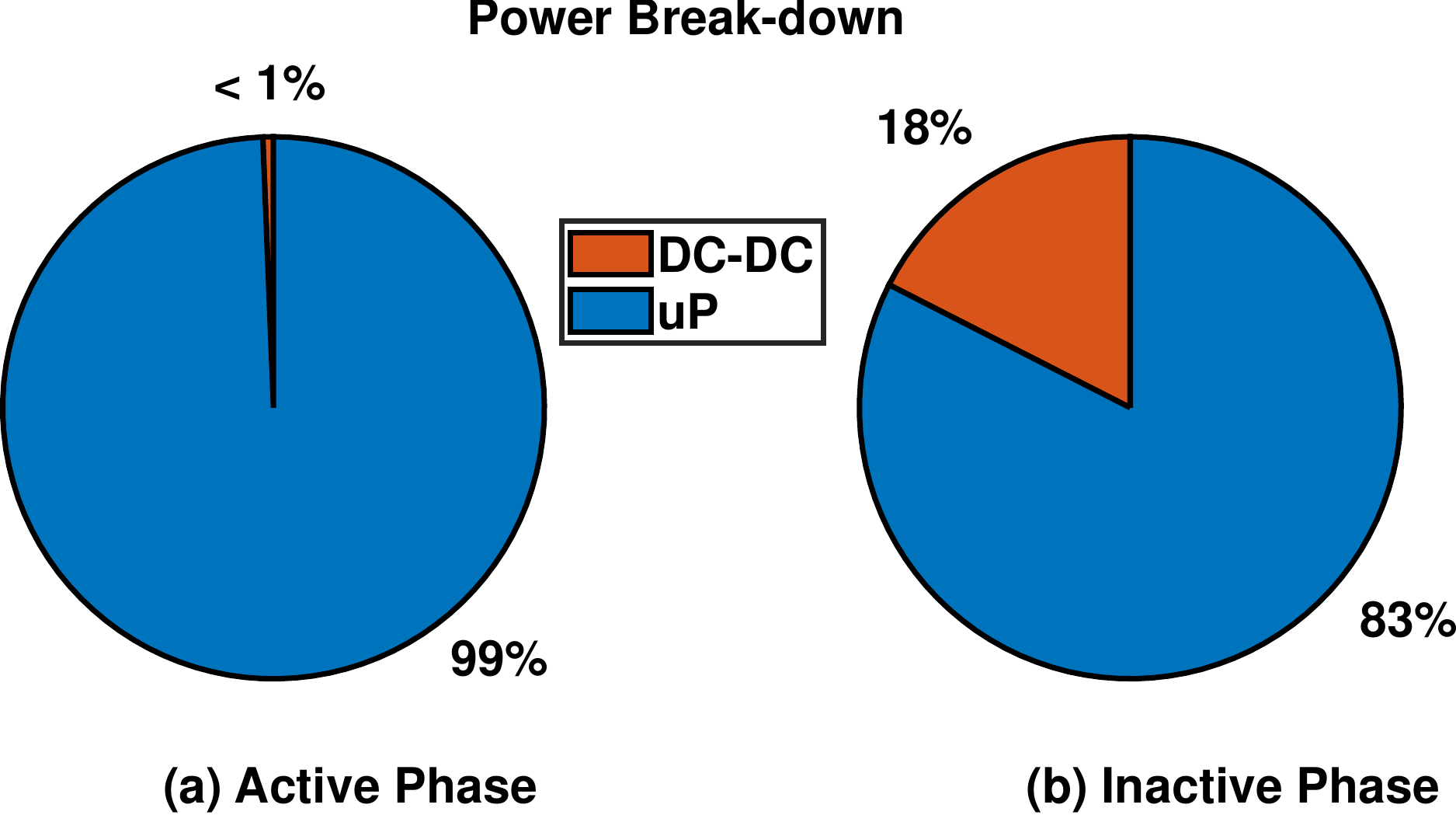}
\caption{Power drawn by DC-DC and uP sub-block in (a) active and (b) inactive phase. In the active period, contribution of DC-DC converter to the overall power consumption is only $0.6\%$.}
\label{energy_break}
\end{figure}

Considering the efficiency factor of DC-DC converter into account, the overall system combining the uP and the buck draws $744\mu$W power in the active and $12\mu$W power in the inactive period. Power break-down for DC-DC and uP sub-block are shown in Fig. \ref{energy_break}. It is apparent that the contribution of DC-DC to the overall energy consumption in the active period is insignificant. Further, if we consider a duty cycled operation where the inactive periods are fixed by the NASA bearings sampling timings and active time by single BL execution time, the overall system power calculated based on the Eq. (\ref{eqn6b}) is similar to the inactive power of the system ($12\mu$W). The overall system power is limited by the inactive power due to the high duty cycle ratio ($114\mu$s: $600$s). This demands the improvement of the efficiency and lowering the inactive power of the system to further increase the longevity of the battery operated system.

Table \ref{tab3} highlights the comparison of our work with other published works on NASA bearing data. None of the work supports online training except for our work. Most of the works used $1$ or $2$ faulty bearings for multi-class defect classification and achieved $97.5$\% detection accuracy. Whereas we achieved 100\% detection accuracy on both healthy and faulty bearing.

Several works on run time configurable system based on the output quality are captured in Table \ref{tab3app}. Energy saving in the table largely depends on the algorithm, input data, level of approximation, and approximation among different subsystems. Approximation in three subsystems- sensors, memory and processor are introduced in \cite{Raha2017}. Signal sub-sampling at the sensor (camera) reduces the data to be stored and data to be processed, and has immense impact on energy saving. Likewise, \cite{Alvarez2017} tunes threshold dynamically and reduces the pixels to be processed at the subsequent stages and achieved $2.85$X energy improvement. Most of the approximation works in the literature are on multimedia application and show their results on existing FPGA board. Whereas, we implement the PdM system in ASIC which can be powered by a single battery cell. To the best of our knowledge, this is the first work which discusses the application of approximate computing throughout the lifetime of the machine.

\begin{table*}[t]
\centering
\caption{Comparison with other published works on NASA bearing data}
    \begin{tabular} {|M{2.3cm}|M{1.7cm}|M{3.5cm}|M{2.7cm}|M{2.7cm}|M{1.5cm}|}\hline
         &\cite{Martinez-Rego2011} &\cite{Tobon-Mejia2012} &\cite{Yu2012} &\cite{Yacout2012} &This Work \\ \hline
         Algorithm  &SVM-OCC  &WPD and MoG-HMM  &LNPP &LAD &ELM-OCC \\ \hline
        Online Learning	&No &No &No &No &Yes\\ \hline
        \#Bearing Used &1 &12 &1 &2 &12\\ \hline
        Classification/RUL &OCC &RUL &Multi-class classification &Multi-class classification &OCC\\ \hline
        Detection Accuracy &Noise analysis & Estimate remaining useful life &97.22\% &97.5\% &100\%\\ \hline
    \end{tabular}
\label{tab3}
\end{table*}

\begin{table*}[t]
\centering
\caption{Comparison with other published works on Approximate Computing}
    \begin{tabular} {|M{2.4cm}|M{2.4cm}|M{3.4cm}|M{4cm}|M{2.5cm}|}\hline
         &~\cite{Raha2015} &\cite{Raha2017} &\cite{Alvarez2017} &This Work \\ \hline
         Approximation level  &Circuit  &System   &system &Software \\ \hline
                  Approach & Reconfigurable Adder \& subtractor  & adaptive image subsampling, DRAM refresh rate \& Computation skipping & Adaptive threshold, keypoint and keypoint description reduction & Dynamic Network Scaling \\ \hline
        Implementation details \& Tech. & FPGA & FPGA  & ASIC, $40$nm CMOS & ASIC, $65$nm CMOS\\ \hline
         Application Area  &Multimedia  &Smart Camera  &IoT Vision &PdM \\ \hline
           Energy Saving  &$1.6$X   &$7.5$X   &$2.85$X \& $5.7$X (Voltage scaling) &$8.95$X \\ \hline
    \end{tabular}
\label{tab3app}
\end{table*}

\section{Case study: Seizure Detection}
\label{disc}
In order to show that the proposed ADEPOS algorithm is generic, we choose EEG dataset for seizure detection from the UPenn and Mayo Clinic’s Seizure Detection Challenge database~\cite{eegdataset}. The database contains $12$ EEG datasets among which $8$ of those are from human and $4$ are from canine subject. Each dataset has $1-$second EEG clips labeled ''Ictal" for seizure data segments and ''Interictal" for non-seizure data segments. The canine data are recorded continuously at $400$Hz deploying $16$ implanted subdural electrodes. Whereas, the human data are from patients undergoing evaluation for epilepsy surgery sampled at $500$Hz or $5000$Hz and have varying numbers of electrodes. Despite the dataset has both training and testing data, we choose the training data for training and testing the ADEPOS algorithm because the training data from the dataset are labeled.

\begin{figure}[t]
\centering
\includegraphics[scale=0.345]{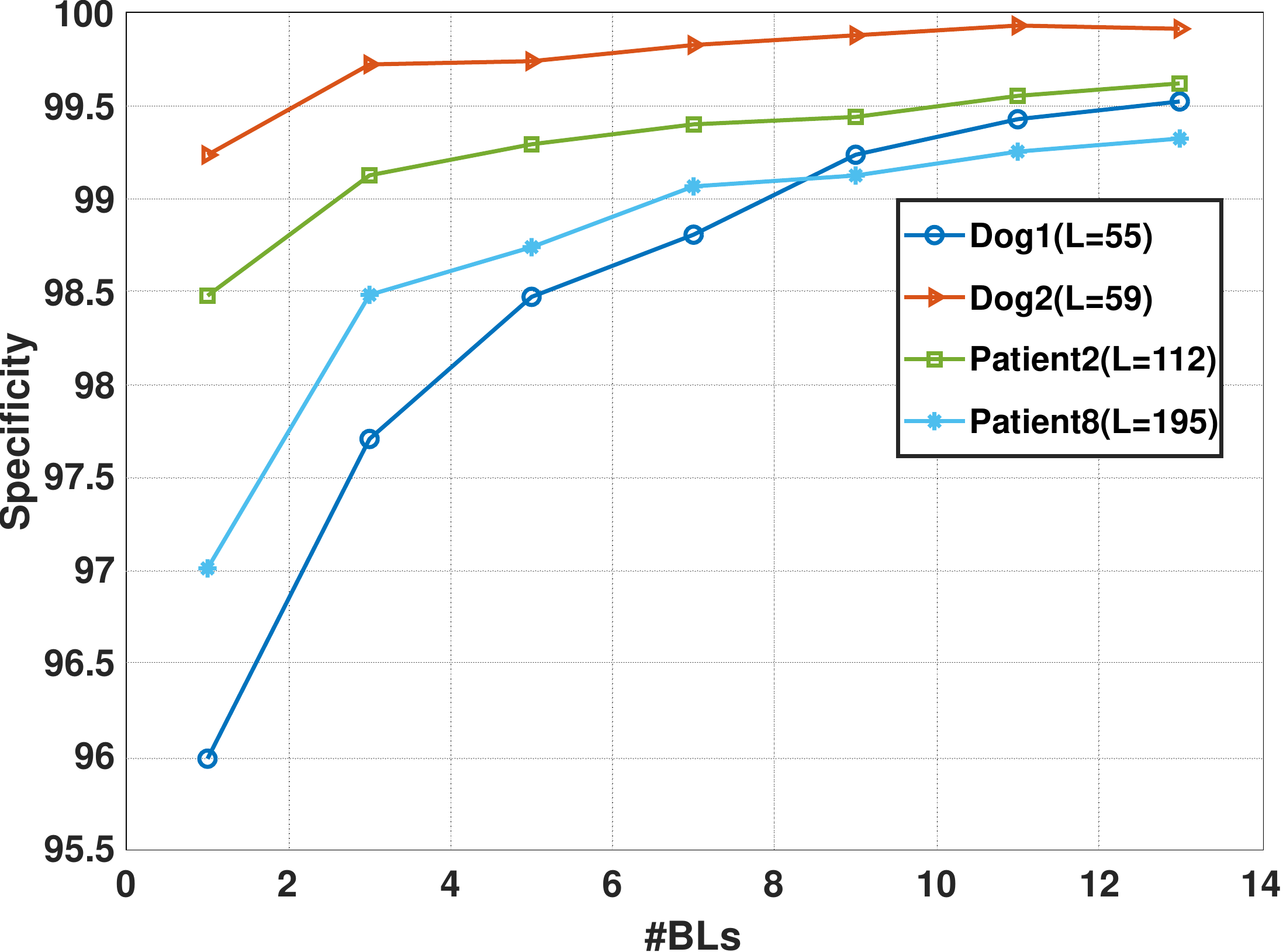}
\caption{Specificity curves of EEG dataset~\cite{eegdataset} for different number of BLs in the network. Sensitivity of $100\%$ is achieved for all the datasets.}
\label{specificity}
\end{figure}

We model the seizure detection as a binary classification problem and randomly select	$50$\% interictal and ictal data for training the model and remaining data are used for testing. Although RMS feature is extracted from each EEG clip for training and testing, other statistical features can also be used. In order to show the generalization of ADEPOS algorithm, we choose two canine and two human datasets having $16$ electrodes. The specificity curves of four EEG datasets in Figure \ref{specificity} shows that the specificity increases with a higher number of BLs are ensemble together and exceeds $99.5\%$ when the number of BL is $13$. Moreover, we achieve $100\%$ sensitivity for all the four datasets. The best value of L of each BL for each dataset is found out using a linear search.
\begin{figure}[h!]
\includegraphics[scale=0.36]{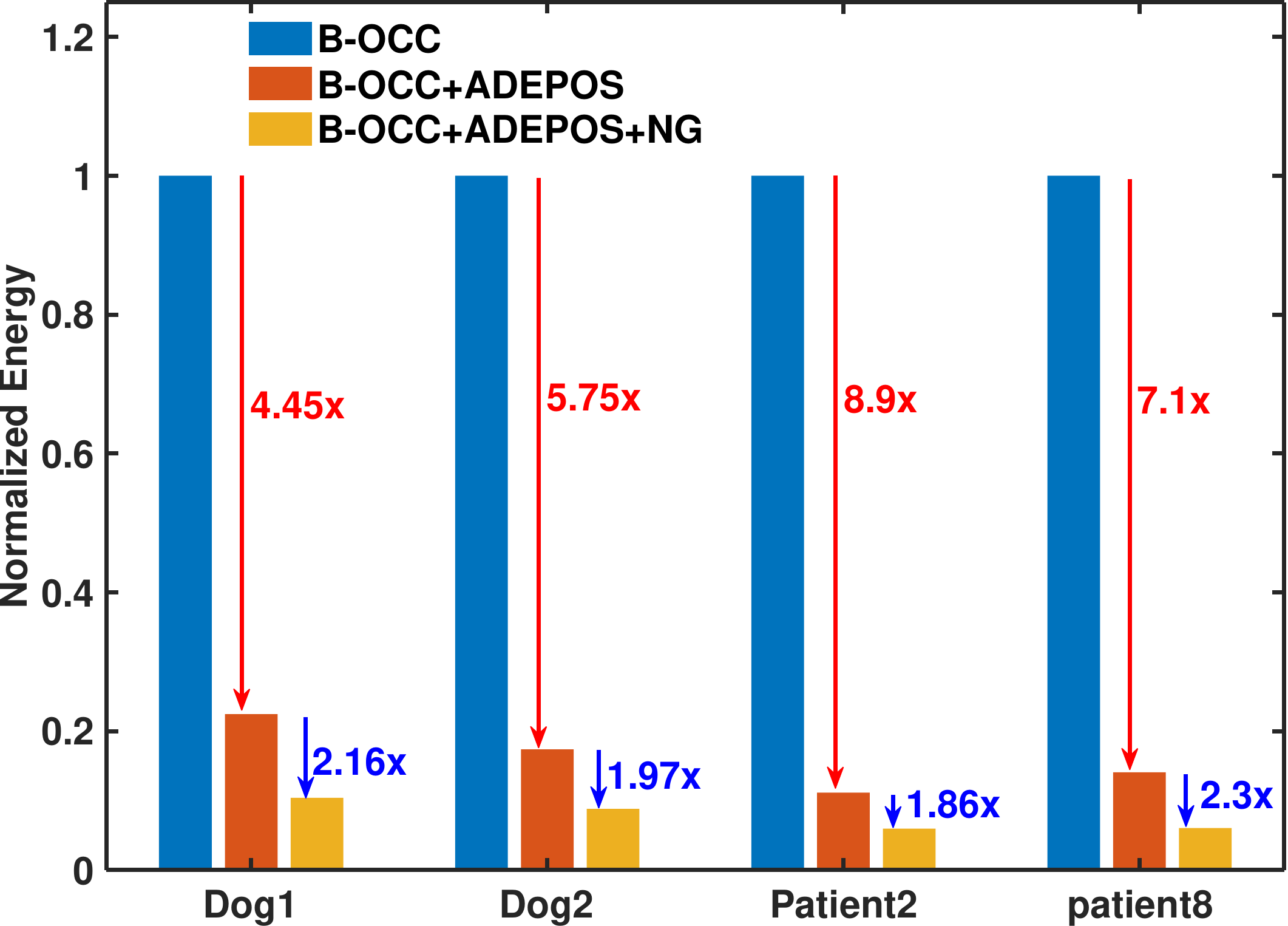}
\caption{Normalized energy saving on EEG dataset deploying ADEPOS and NG algorithm. ADEPOS enables $6.56$X average energy saving over B-OCC. NG further reduces the energy requirement by a factor of $2$.}
\label{eeg_energy}
\end{figure}

\begin{table}[b]
\centering
\caption{Comparison with other published works on EEG dataset}
    \begin{tabular} {|M{1.7cm}|M{1.9cm}|M{1.9cm}|M{1.3cm}|}\hline
          &\cite{Zhang2015} &\cite{Koteshwara2018} &This work \\ \hline
        Features	 & Spectral power and ratios   &Spectral power and ratios  &RMS \\ \hline
        Sensitivity	 &100  &98.2  &100 \\ \hline
        Specificity &99.9  &98.88 &99.93\\ \hline
        Energy saving &1X   &1.53X &13.1X\\ \hline
    \end{tabular}
\label{tab8}
\end{table}

We apply the proposed ADEPOS algorithm on the EEG dataset and find  out the average number of BLs for all the test samples of Dog1, Dog2, Patient2 and Patient8 are $2.92$, $2.26$, $1.45$ and $1.83$ respectively. We have estimated the energy consumption of different algorithms and approaches based on Fig. \ref{energy_bl} and plotted the normalized energy for different datasets in Fig.~\ref{eeg_energy}. It can be seen from Fig. \ref{eeg_energy} that the ADEPOS enables $6.56$X average energy saving on EEG dataset and NG approach further reduces the energy consumption by a factor of $2$. Moreover, compared to standalone B-OCC, NG reduces $69\%$ program memory footprint for the EEG dataset.

Table \ref{tab8} shows the comparison of the proposed work with other works on the EEG dataset. Both~\cite{Zhang2015} and~\cite{Koteshwara2018} are using absolute spectral power, relative spectral power and spectral power ratio as features which are more expensive in terms of computation than the RMS feature. The specificity of the proposed method is $99.93\%$ which is higher than the previously reported specificity on the EEG dataset. Moreover, the proposed method achieves $13.1$X energy saving.

\section{Conclusion}
\label{conclu}
Aiming at industry 4.0, we have proposed to use OPIUM-B OCC for predictive maintenance deployed in machine health monitoring. OPIUM-B OCC not only consumes lesser energy during learning and inference phase but also has a lower memory footprint than existing AE-OCC. To reduce the energy consumption of PdM sensor further, we have suggested to introduce approximate computing adaptively throughout the lifetime of the machine, since at the early stage of the machine, there will be less degradation in the data. We have shown that the proposed ADEPOS and B-OCC enable an average energy saving of $8.95$X without losing any detection accuracy. We have also achieved $13.1$X energy saving on the EEG dataset using ADEPOS and NG method together. Moreover, NG enables $64\%$-$69\%$ reduction of the program memory requirement to execute the proposed ADEPOS algorithm. Finally, the synergy between the processor and DC-DC converter helps to reduce the energy further leveraging voltage scaling. The whole system can be powered by a single lithium-ion battery cell and directly deployed for machine health monitoring.

\appendix[Derivation of Adaptive On Time,~$T_{on}$]
In the following derivation, we use Figure \ref{ripple} and all the notations are annotated in the figure.

Once the latch, L1, is set by the rising edge of the $V_{out}$ comparator output, the capacitor, $C$, start charging through P1 and  R until the voltage across C reaches $\frac{2\cdot V_{batt}}{3}$. Since the initial and final voltages across R are $V_{batt}$ and $\frac{V_{batt}}{3}$ respectively, the average current through R, $I_R$, during $T_{on}$ follows Eq. (\ref{eqnaa}).
\begin{equation}
I_R=\frac{2\cdot V_{batt}}{3\cdot R}
\label{eqnaa}
\end{equation}

\begin{figure}[t]
\includegraphics[scale=0.37]{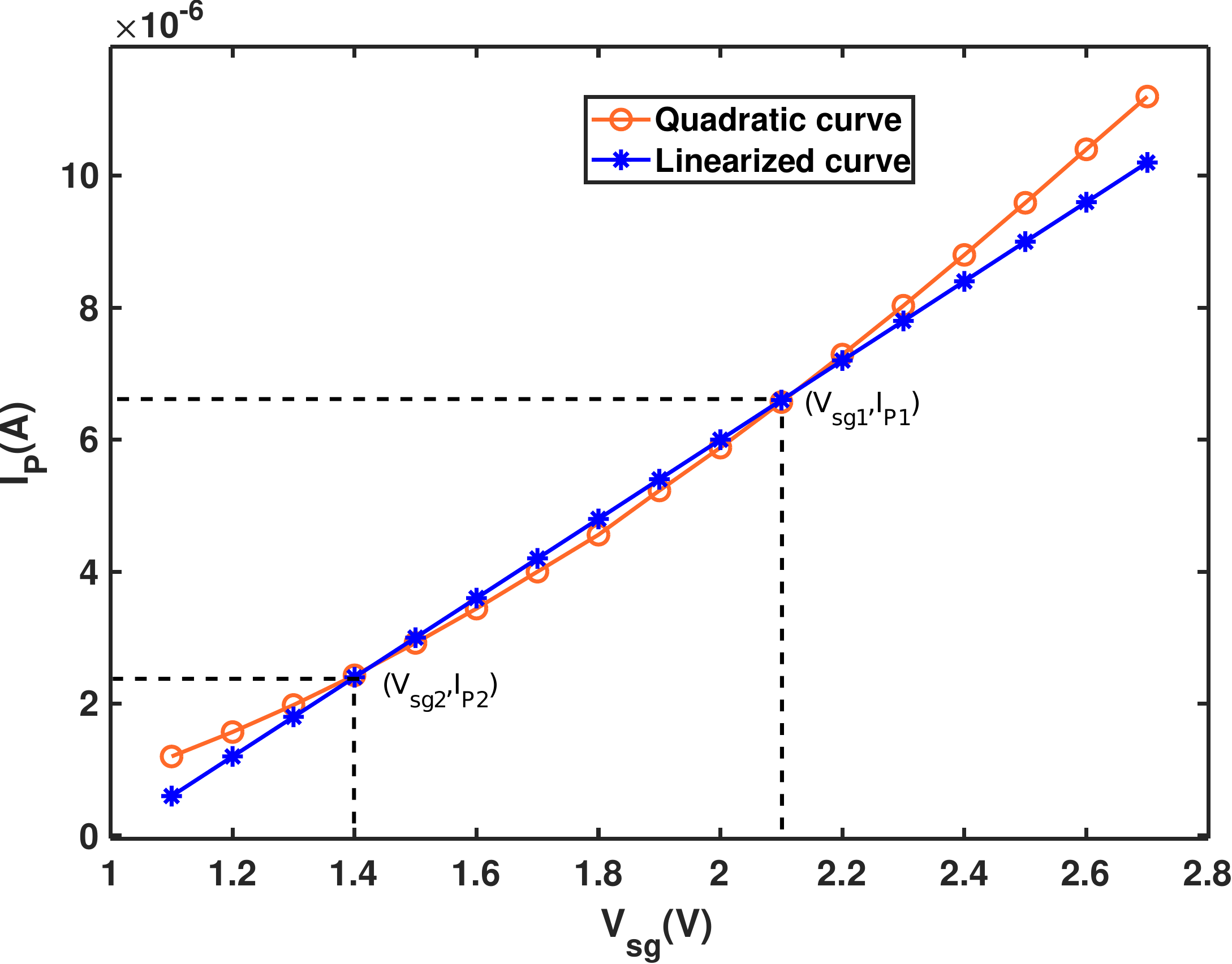}
\caption{Linearization of P1 drain current between $1.1$V to $2.7$V of $V_{sg}$. Error in the derivation of overall charging current of capacitor C due to the approximation is less than $9\%$ at $V_{sg}=1.1$V where the resistor current is dominant.}
\label{linear}
\end{figure}

On the other hand, the current through P1 is quadratic in nature and shown in Eq. (\ref{eqnab}) where k depends on the process and the aspect ratio of P1 and $V_{th}$ denotes the threshold voltage of P1.
\begin{equation}
I_P=k\cdot (V_{sg}-|V_{th}|)^2
\label{eqnab}
\end{equation}
Since the $V_{sg}$ of P1 varies between $1.1$V to $2.7V$ based on the $V_{batt}$ and $V_{out}$ variation, we linearize Eq. (\ref{eqnab}) in this range which follows Eq. (\ref{eqnac}) and is also shown in Fig. \ref{linear}.
\begin{align}
I_P &=k_3\cdot V_{sg}+k_4 \label{eqnac}\\
k_3 &=k\cdot (V_{sg1}+V_{sg2}-2|V_{th}|)  \label{eqnad}\\
k_4 &= I_{P1}-k\cdot V_{sg1}\cdot (V_{sg1}+V_{sg2}-2|V_{th}|)  \label{eqnae}
\end{align}
\begin{table}[h!]
\centering
\caption{PVT variation of $T_{on}$}
    \begin{tabular} {|M{1cm}|M{1cm}|M{1cm}|M{1cm}|}\hline
    \multicolumn{2}{|c|}{ $V_{batt}=3.3$V} &\multicolumn{2}{|c|}{$V_{batt}=2.5$V}\\ \hline
        P/T  &$T_{on}(nS)$ &P/T &$T_{on}(nS)$ \\ \hline
        tt/27	 &317   &tt/27  &416 \\ \hline
        ss/125 &393   &ss/125 &480\\ \hline
        ff/-40 &267   &ff/-40 &352\\ \hline
    \end{tabular}
\label{tab7}
\end{table}

Hence, the adaptive on time, $T_{on}$, can be derived as follows
\begin{align}
T_{on}&= \frac{k1 \cdot V_{batt}}{(k2 \cdot V_{batt}- k3 \cdot V_{out})+k4} \label{eqnaf} \\
k1 & =\frac{2 C}{3} \label{eqnag} \\
k2 & =k_3+\frac{2}{3R} \label{eqnag}
\end{align}

Table \ref{tab7} captures the simulation results of PVT variation of $T_{on}$ at $V_{batt}=3.3$V and $2.5$V. At lower $V_{batt}$, $T_{on}$ increases which in turn reduces output voltage variation. Across process and temperature $T_{on}$ shows $-23\%$ to $19\%$ variation which can be corrected by trimming R and P1.


%



\section*{Acknowledgment}
This work was conducted within the Delta-NTU Corporate Lab for Cyber-physical Systems with funding support from Delta Electronics Inc. and the National Research Foundation (NRF) Singapore under the Corp Lab @University scheme.
\ifCLASSOPTIONcaptionsoff
  \newpage
\fi



%


\bibliographystyle{ieeetr}
\footnotesize
\bibliography{adepos}

%










\end{document}